\pdfoutput=1
\documentclass[11pt]{article}

\usepackage{latex/acl}

\usepackage{times}
\usepackage{latexsym}

\usepackage[T1]{fontenc}
\usepackage[utf8]{inputenc}

\usepackage{microtype}

\usepackage{inconsolata}

\usepackage{times}  % DO NOT CHANGE THIS
\usepackage{helvet}  % DO NOT CHANGE THIS
\usepackage{courier}  % DO NOT CHANGE THIS
\usepackage{natbib}  % DO NOT CHANGE THIS AND DO NOT - ANY OPTIONS TO IT
\usepackage{caption} % DO NOT CHANGE THIS AND DO NOT - ANY OPTIONS TO IT
\usepackage{bm}
\usepackage{helvet}
\usepackage{courier}
\usepackage{graphicx}
\usepackage{bbding}    
 \usepackage{amssymb}

\usepackage{soul}
\usepackage{soul}
\usepackage[utf8]{inputenc}
\usepackage{graphicx}
\usepackage{amsmath}
\usepackage{amsthm}
\usepackage{booktabs}
\usepackage{algorithm}
\usepackage{algorithmic}
\usepackage{hyperref}
\usepackage{soul}
\usepackage[utf8]{inputenc}
\usepackage{graphicx}
\usepackage{amsmath}
\usepackage{amsthm}
\usepackage{booktabs}
\usepackage{algorithm}
\usepackage{algorithmic}
\usepackage[switch]{lineno}
\usepackage{color}
\usepackage{array}
\usepackage{booktabs}
\usepackage{multirow}
\usepackage{bm}
\usepackage{enumitem}
\usepackage{latexsym}
\usepackage{array}
\usepackage{bm}
\usepackage{helvet}
\usepackage{courier}
\usepackage{graphicx}
\usepackage{soul}
\usepackage{soul}
\usepackage[utf8]{inputenc}
\usepackage{graphicx}
\usepackage{amsmath}
\usepackage{amsthm}
\usepackage{booktabs}
\usepackage{algorithm}
\usepackage{algorithmic}
\usepackage{soul}
\usepackage[utf8]{inputenc}
\usepackage{graphicx}
\usepackage{amsmath}
\usepackage{amsthm}
\usepackage{booktabs}
\usepackage{algorithm}
\usepackage{algorithmic}
\usepackage[switch]{lineno}
\usepackage{color}
\usepackage{array}
\usepackage{booktabs}
\usepackage{multirow}
\usepackage{bm}
\usepackage{enumitem}
\usepackage{mathrsfs}
\usepackage{booktabs}
\usepackage{pifont}
\usepackage{utfsym}
\usepackage{fontawesome}

\usepackage{hyperref}       % hyperlinks
\usepackage{url}            % simple URL typesetting
\usepackage{booktabs}       % professional-quality tables
\usepackage{amsfonts}       % blackboard math symbols
\usepackage{nicefrac}       % compact symbols for 1/2, etc.
\usepackage{microtype}      % microtypography
\usepackage{graphicx} 
\usepackage{subcaption} 
\usepackage{amsmath}
\usepackage{mathtools}
\usepackage{algorithm}
\usepackage{newfloat}
\usepackage{listings}
\newtheoremstyle{mystyle}  % 主定理样式
  {6pt}   % 上方间距
  {6pt}   % 下方间距 
  {\itshape}  % 正文字体
  {}     % 缩进
  {\bfseries} % 标题字体
  {.}    % 标题后标点
  { }    % 标题后间距
  {}     % 自定义头部

\newtheoremstyle{defstyle}  % 定义专用样式
  {6pt}
  {6pt}
  {\normalfont}  % 正体字体
  {}
  {\bfseries}
  {.}
  { }
  {}

\theoremstyle{mystyle}
\theoremstyle{defstyle}
\usepackage{etoolbox}
\patchcmd{\proof}{\itshape}{\itshape\leavevmode}{}{}
\patchcmd{\proof}{\pushQED{\qed}}{$\hspace*{2em}\pushQED{\qed}$}{}{}

\title{S\textsuperscript{2}Sent: Nested Selectivity Aware Sentence Representation Learning}

\author{
\textbf{Jianxiang Zang}$^{1}$, 
\textbf{Nijia Mo}$^{2}$, 
\textbf{Yongda Wei}$^{2}$, 
\textbf{Meiling Ning}$^{3}$\textbf{,}
\textbf{Hui Liu}$^{2}$\thanks{Corresponding author.}\textbf{,} 
\\
\normalsize{$^{1}$ Fudan University},
\normalsize{$^{2}$ Shanghai University of International Business and Economics} \\  % 修正行
\normalsize{$^{3}$ Beijing University of Posts and Telecommunications},\\
\normalsize{\texttt{jxzang25@m.fudan.edu.cn, liuh@suibe.edu.cn}}  % 删除行尾的 \\
}
\begin{document}
\maketitle

\begin{abstract}
The combination of Transformer-based encoders with contrastive learning represents the current mainstream paradigm for sentence representation learning. This paradigm is typically based on the hidden states of the last Transformer block of the encoder. However, within Transformer-based encoders, different blocks exhibit varying degrees of semantic perception ability. From the perspective of interpretability, the semantic perception potential of knowledge neurons is modulated by stimuli, thus rational cross-block representation fusion is a direction worth optimizing. To balance the semantic redundancy and loss across block fusion, we propose a sentence representation selection mechanism S\textsuperscript{2}Sent, which integrates a parameterized nested selector downstream of the Transformer-based encoder. This selector performs spatial selection (SS) and nested frequency selection (FS) from a modular perspective. The SS innovatively employs a spatial squeeze based self-gating mechanism to obtain adaptive weights, which not only achieves fusion with low information redundancy but also captures the dependencies between embedding features. The nested FS replaces GAP with different DCT basis functions to achieve spatial squeeze with low semantic loss. Extensive experiments have demonstrated that S\textsuperscript{2}Sent achieves significant improvements over baseline methods with negligible additional parameters and inference latency, while highlighting high integrability and scalability.
\end{abstract}

\section{Introduction}
The integration of Transformer-based encoders and contrastive learning is still the mainstream paradigm for sentence representation learning \cite{devlin2019bert,liu2019roberta,gao2021simcse,jiang2022promptbert,warner2024smarter}. BERT-based embeddings have been successfully deployed in large-scale retrieval systems~\cite{zhu2024longembed}, achieving high-throughput semantic matching under sub-millisecond latency constraints \cite{gao2021simcse}. In real-time interactive scenarios, sequence recommendation architectures incorporating sentence representations significantly improve click-through rates through contrastive learning strategies \cite{zheng2024harnessing,zang2025mitigating}. For retrieval-augmented generation systems, dynamically distilled sentence embedding optimization substantially reduces computational overhead in real-time semantic matching~\cite{zang2023extract,zang2023improving,zhao2024retrieval}. Enhancing the separation of sentences with distinct semantics through contrastive learning is the main research trend for improving sentence representation learning. 

However, an overlooked issue is that these methods typically rely on the hidden states from the last Transformer block of the encoder, with different blocks of the encoder exhibiting varying levels of semantic sensitivity. There is also some work advocates linear fusion of hidden states from different blocks or equal consideration of them during training~\cite{su2021whitening,gao2021simcse,zhuo2023whitenedcse}. In the field of neuroscience, the semantic potential of neurons is modulated by stimuli, which is not static and is related to the contrast of the stimulus~\cite{nelson1978orientation}: the lower the contrast, the stronger the effective semantic perception~\cite{sceniak1999contrast}. From the perspective of interpretability, the lower-level blocks of Transformer-based encoder models encode the basic and structural semantic information (for example, part-of-speech tags), while the later layers encode more “semantic” information~\cite{tenney2019bert}. Recently, related studies have found that “knowledge neurons” are distributed in the top several layers of pre-trained encoder models~\cite{dai2022knowledge}.
Treating the hidden states from different blocks equally is insufficient to model this characteristic and difficult to unleash the semantic perception potential of different blocks. Inspired by this, we propose a novel selectivity aware cross-block fusion mechanism \textbf{S\textsuperscript{2}Sent}, which incorporates a parameterized nested selector downstream of Transformer-based encoder. This selector performs a \textbf{S}patial \textbf{S}election (\textbf{SS}) nested with a \textbf{F}requency \textbf{S}election (\textbf{FS}). Through these selections, S\textsuperscript{2}Sent modulates feature extraction at different scales within the encoder, fusing a more potential sentence representation while reducing downstream redundancy and semantic information loss.

The SS module is a selective multi-scale adaptive fusion mechanism that employs a novel spatial squeeze based self-gating mechanism. It employs a squeeze-excitation mechanism to obtain feature representations of hidden states, using them as adaptive weights to fuse the hidden states of each block. This method of determining adaptive weights avoids the information redundancy caused by conflicts between the previous self-gating~\cite{rei2019jointly,stacey2022supervising} mechanisms and the residual connections in Transformers. It also models dependencies between features that were previously unexplored in Transformers, highlighting features that are crucial for sentence semantics. The FS module, nested within the SS, is a multispectral selective spatial squeeze that leverages the property that the discrete cosine transform (DCT) is an extension of global average pooling (GAP), by selecting low-frequency DCT basis functions for each feature component to be encoded as a scalar. Compared to GAP-based spatial squeeze~\cite{hu2018squeeze,li2019selective,li2023large}, the FS module largely preserves the expressive power of the feature channels during spatial squeeze.

%SS模块是一种选择性的多尺度自适应融合机制，它用一种新颖的squeeze-excitation机制去获取隐藏状态的特征表示，作为自适应权重对每个块的隐藏状态进行融合。这种自适应权重的确定方法避免了之前self-gating~\cite{rei2019jointly,stacey2022supervising}机制与Transformer中残差连接冲突造成的信息冗余。同时建模了Transformer中以前未探索的特征之间的依赖关系，突出了对句子语义至关重要的特征。
%嵌套在SS之内的FS模块是一个多光谱选择性空间squeeze，利用离散余弦变化(DCT)是全局平均池化(GAP)的拓展这一特性，通过为每个特征分量压缩选择低频DCT基函数来编码为一个标量。相对于基于GAP的空间压缩~\cite{hu2018squeeze,li2019selective,li2023large}，FS在空间压缩上很大程度上维护了特征通道的表达能力。

We evaluate four baseline methods using BERT~\cite{devlin2019bert} and RoBERTa~\cite{liu2019roberta} as backbones, and perform unsupervised sentence embedding training with SimCSE~\cite{gao2021simcse} and PromptCSE~\cite{jiang2022promptbert}. We analyze the impact of S\textsuperscript{2}Sent on these baselines across seven STS tasks. Extensive experimental results demonstrate that S²Sent is compatible with all Transformer-based encoders and can significantly enhance the sentence representation capability of the backbone with only negligible increases in additional parameters and inference latency. The ablation studies further emphasize the contributions of these two selection mechanisms. Through the discussion of hyperparameters, we were pleasantly surprised to find that S\textsuperscript{2}Sent does not rely on the selection of hyperparameters such as the number of Transformer blocks or the frequency components, demonstrating strong integrability and scalability. Our primary contributions are highlighted as follows:

\begin{itemize}
    \item We propose a novel sentence representation mechanism, S\textsuperscript{2}Sent. S\textsuperscript{2}Sent achieves efficient sentence representation fusion across Transformer blocks by leveraging Spatial Selection (SS) and nested Frequency Selection (FS).
    \item SS introduces a novel self-gating mechanism that assigns spatially squeezed feature representations as adaptive weights to achieve representation fusion. This method not only reduces information redundancy but also models the dependencies between embedding features.
    \item FS selects different low-frequency DCT basis functions to weight and squeeze feature parts, thereby reducing the loss associated with feature channel scalarization.
    \item S²Sent is compatible with all Transformer-based encoders, and the additional parameters and inference latency it introduces are negligible compared to the backbone network. Moreover, the optimization it brings does not rely on the choice of hyperparameters, demonstrating its high integrability and scalability.
\end{itemize}

\begin{figure*}[h]
\centering
\includegraphics[width=1\textwidth]{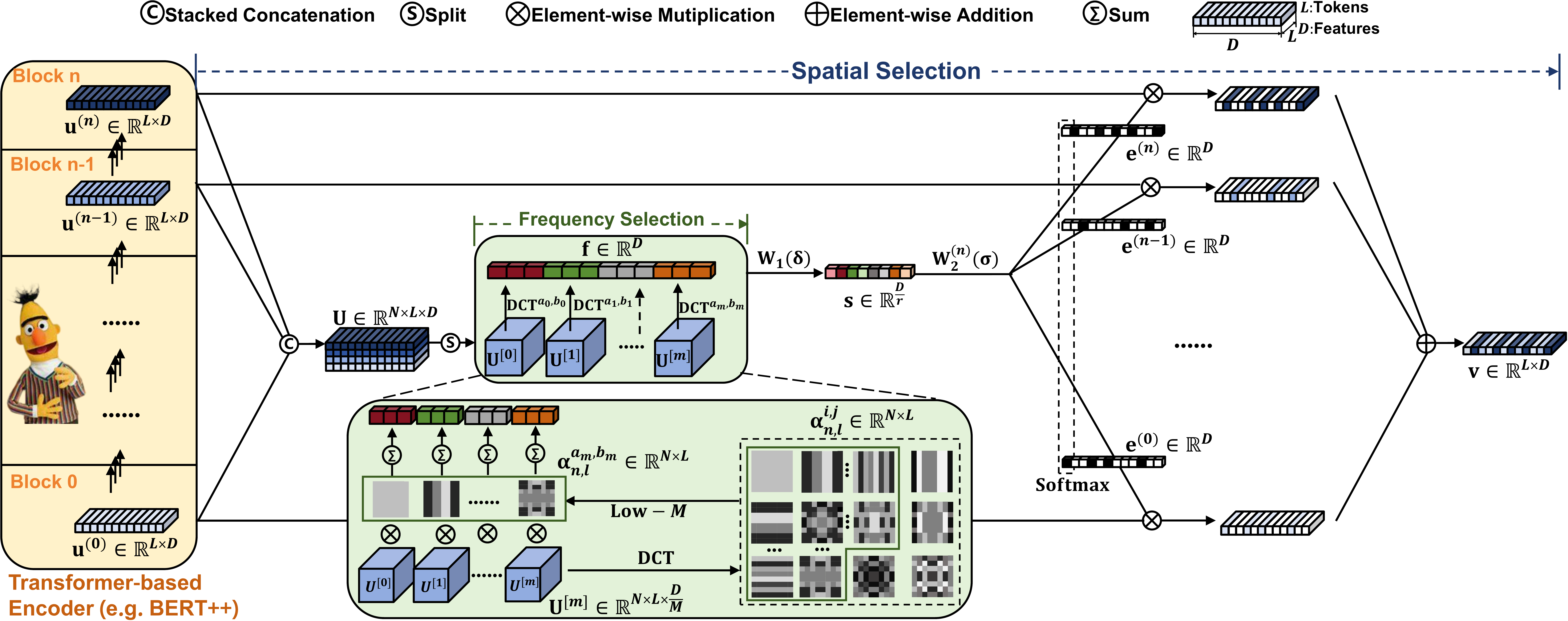}
\caption{Overview of S\textsuperscript{2}Sent. S\textsuperscript{2}Sent builds a parameterized nested selector downstream of Transformer-based encoder, which performs a Spatial Selection (SS) nested with Frequency Selection (FS).
}\label{fig.s2a}
\end{figure*}

\section{Related Work}

\subsection{Sentence Representation Learning}
%最近在大模型的风口下，基于LLM的句子表征任务仍然难以取得良好的表现，大模型的训练都是基于文本续写形式来训练的，损失函数主要预测下一个Token是否准确，而不是判断整个句子表征的好坏。
Amidst the recent surge in large language models, the task of sentence representation based on LLMs continues to struggle to achieve satisfactory performance. These models are predominantly trained on text continuation tasks, where the loss function primarily focuses on the accuracy of predicting the next token, rather than evaluating the quality of the overall sentence representation~\cite{zhu2023large,wang2024improving}.
Accordingly, Transformer-based pre-trained encoders are still the mainstream sentence representation backbone, adopting an end-to-end approach~\cite{devlin2019bert,liu2019roberta,he2020deberta}. However, these methods encode sentences out-of-the-box, leading to issues such as anisotropy in the representation space~\cite{ethayarajh2019contextual,zhang2020unsupervised,zang2025compression}. Some efforts have been made to transform the representation space into a smooth and isotropic space~\cite{li2020sentence,su2021whitening,zhuo2023whitenedcse}. Contrastive learning methods, on the other hand, aim to tightly combine semantically similar examples while further separating different examples~\cite{gao2021simcse,zhang2022unsupervised,jiang2022promptbert}, and have emerged as mainstream learning approaches. However, a neglected perspective is that these methods are based on the hidden states from the last Transformer block of the encoder. There is also some work that considers the linear fusion of hidden states from different blocks or equal consideration of them during training~\cite{su2021whitening,gao2021simcse,zhuo2023whitenedcse}. 

However, treating them equally cannot model the adaptive semantic perception of different blocks, and we will demonstrate this in the discussion of the motivation for S\textsuperscript{2}Sent.

%最近，基于Transformer的预训练编码模型已成为句子表示的主干，采用端到端的方法~\cite{devlin2019bert,liu2019roberta,he2020deberta}。然而，这些方法开箱即编码句子，导致表示空间中的各向异性等问题~\cite{ethayarajh2019contextual,zhang2020unsupervised}。一些努力已经被投入到将表示空间转换为平滑且各向同性的空间，例如归一化流变换~\cite{li2020sentence}和白化变换~\cite{su2021whitening,zhuo2023whitenedcse}。另一方面，对比学习方法旨在紧密地结合语义相似的样本，同时进一步分离不同的样本~\cite{gao2021simcse,zhang2022unsupervised,jiang2022promptbert}，并已成为主流的学习方式。还有大量相关工作考虑联合建模不同块的隐藏状态。然而，这些方法通常统一地对待每个块或线性融合不同块的表示~\cite{su2021whitening}。这意味着在微调和反向传播期间，不同尺度上的特征提取趋于收敛到相同的程度，限制了在不同尺度上提取特征的能力。

\subsection{Self-Gate for Composing Sentence Representation}

The difficulty of learning to construct sentence representations in end-to-end systems lies in the fact that natural language is highly compositional. Related studies have \cite{rei2019jointly,zang2024explanation} proposed a self-gate based sentence representation learning architecture. This architecture first activates relevance scores for each token, which can be seen as adaptive weights. The attention weights are then fused with the token sequence embeddings to form the final sentence representation. This self-gate mechanism essentially functions as a sequence tagging system, guiding the encoder to identify important domains in the sentence. Related work has applied this ~\cite{stacey2022supervising} to downstream Transformer-based encoders. However, these methods activate token representations and fusion mechanisms that are essentially consistent with the feed-forward neural networks and residual connections in the Transformer, leading to information redundancy. 

In contrast, we advocate for spatial squeeze to obtain sentence feature representations as adaptive weights for the self-gating mechanism in SS module. This not only achieves low-redundancy fusion but also models feature dependencies that were not present in the Transformer.

%学习在端到端系统中构建句子表示的困难在于自然语言是高度组合的，而且特定于任务的注释数据集的大小往往有限。\cite{rei2019jointly}提出了一种联合学习标记句子和标记的体系结构学习更好的句子表示。该体系首先激活每个token的相关性得分，这些分数可以视作自注意机制的自适应权重。将注意力权重与token序列的embedding进行融合以组合成最后的句子表示。这种self-gate的机制本质为一个序列标记系统，指导编码器识别句子中重要的领域。相关工作将这种结构~\cite{stacey2022supervising}应用于Transformer based encoder的下游。然而这些方法激活token的表示和融合机制分别与Transformer中的前馈神经网络和残差连接本质上一致，这无疑造成了信息的冗余。我们则主张进行空间Squeeze获得句子特征表示作为self-gate的自适应权重。这不但分配了自适应权重的同时进行了Transformer中不曾有的特征间依赖建模。

\subsection{Spatial Squeeze for Feature Representation}

Channel attention mechanisms have demonstrated exceptional performance in computer vision tasks~\cite{hu2018squeeze,woo2018cbam,hou2020strip}. The squeeze-and-excitation network (SE-Net)~\cite{hu2018squeeze} pioneered channel attention by effectively capturing interdependencies among channels through spatial squeeze of feature maps. Subsequent studies, such as CBAM~\cite{woo2018cbam}, SK~\cite{li2019selective}, LSK~\cite{li2023large}, SFA~\cite{zang2024modeling}, DDualSE~\cite{mo2025ddualse}, expanded on this idea by incorporating various spatial encoding or attention mechanisms.
Inspired by these developments, we propose to leverage downstream attention to model dependencies among semantic embedding features and enhance semantic representation. However, the core spatial squeeze techniques in these approaches still rely on GAP and simply averaging the feature channels for scalarization inevitably leads to significant semantic loss.

In contrast, we consider adopting the DCT extension of GAP, utilizing more high-frequency components to achieve spatial squeeze in FS module.

%通道级注意机制在计算机视觉领域~\cite{hu2018squeeze,hu2018gather,woo2018cbam,hou2020strip}展现出了出色的性能。Squeeze-and-Excitation（SE-Net）~\cite{hu2018squeeze}通过GAP对特征图进行空间压缩进行特征表示，有效地构建了通道之间的相互依赖关系，从而开创了通道注意力的先河。随后大量的研究，如CBAM~\cite{woo2018cbam}、GALA~\cite{linsley2019learning}、TA~\cite{misra2021rotate}通过采用不同的空间编码或注意机制扩展了这一思想。受此启发，我们首次利用下游注意力建模语义嵌入特征之间的依赖以增强语义表示。然而其中的核心空间压缩均是保留了GAP，我们在Motivation2证明了GAP仅相当于DCT的最低频率分量，这意味着许多其他潜在有用的高频成分未被探索。

\section{Methodology}

%如图~\ref{fig.s2a}所示，S\textsuperscript{2}Sent是一个基于Transformer-based Encoder的下游参数化映射$\mathcal{F}_{\rm S\textsuperscript{2}Sent}(\mathbf{\theta}):\{\mathbf{u}^{<n>}\}_{n=0}^{N-1}\xrightarrow{\mathbf{v}},\mathbf{\theta}=\{\mathbf{W}^{0},\mathbf{W}^1,\cdots,\mathbf{W}^{(n)}\}$。其中$\mathbf{\theta}$代表了S\textsuperscript{2}Sent中$n+1$个可训练参数，$n$代表了Transformer-based Encoder中参与语义表示的Transformer块。$\{\mathbf{u}^{<n>}\},\mathbf{u} \in \mathbb{R}^{L \times D}$分别是Transformer block中每一层的表示以及最后的语义表示。在文章中,$\mathcal{F}[\mathbf{\theta}]$代表有可训练参数的映射，$\mathcal{F}$代表无可训练参数映射。

%S\textsuperscript{2}Sent针对于Transformer based encoder句子表示任务，旨在激发block级别微调收敛的潜力，并在下游以最小的语义损耗优化出唯一的句子表示

In this work, we propose a novel sentence representation method, S\textsuperscript{2}Sent. It is essentially a nested selectivity aware representation fusion mechanism across Transformer blocks, where this sophisticated nested selection mechanism minimizes semantic redundancy and loss to the greatest extent possible. As illustrated in Figure~\ref{fig.s2a}, S\textsuperscript{2}Sent builds a downstream parameterized selector based on the Transformer-based encoder, performing a tensor mapping: \(\{\mathbf{u}^{(n)}\}_{n=0}^{N-1} \rightarrow \mathbf{v}\). Here, \(\mathbf{u}^{(n)} \in \mathbb{R}^{L \times D}\) denotes the hidden states from each Transformer block, and \(\mathbf{v} \in \mathbb{R}^{L \times D}\) denotes the final sentence representation after optimization by the selector. From a modular perspective, S\textsuperscript{2}Sent performs a Spatial Selection nested with Frequency Selection. The symbols in this paper are explained in Table~\ref{tab.symbol}.

\begin{table}[t]
\scriptsize
\setlength{\tabcolsep}{2.2 pt}
\renewcommand{\arraystretch}{1.3} 
\centering
\begin{tabular}{cc|cc}
\hline
\textbf{Symbol} & \textbf{Interpretation}   & \textbf{Symbol} & \textbf{Interpretation}   \\ \hline
$\mathbf{W}$        & Trainable parameters          & $\cdot_{n,l}$   & Spatial Descriptor        \\ 
$N$             & Transformer blocks number & $\cdot_{d}$     & Feature Descriptor        \\ 
$L$             & Sentence Length           & $<;>$           & Dimensional Concatenation \\ 
$D$             & Embedding dimension       & $[;]$           & Stacked Concatenation     \\ 
$\delta$        & Tanh Fuction              & $\sigma$        & Sigmoid Fuction           \\ \hline
\end{tabular}
\caption{Symbol description sheet}\label{tab.symbol}
\end{table}

\subsection{Spatial Selection}

\subsubsection{Spatial Squeeze based Self-Gate for Adaptive Weights}
%实现跨block的句子表征融合的关键是在于自适应权重的计算，self-gate机制是实现自适应的标准手段。传统的self-gate方法通过获得token级别的激活作为自适应权重，指导编码器关注句子中重要的tokens，如Figure~\ref{fig.sg1}所示。然而这些方法激活token的表示和融合机制分别与Transformer中的前馈神经网络和残差连接本质上一致，这无疑造成了信息的冗余。因此我们主张采用squeeze-excitation的理念，将表征空间squeeze至特征级别表示，建模特征级别的依赖。如Figure~\ref{fig.sg2}所示，以激活压缩后的特征表示作为自适应权重能够指导关注对句子表征有意义的特征，这是Transformer未涉及的建模角度。毫无疑问，这种低信息冗余的self-gate机制是作为句子表征自适应融合权重的首要选择。

\begin{figure}[t]
    \centering
    \begin{minipage}[t]{0.9\linewidth}
        \centering
        \includegraphics[width=\linewidth]{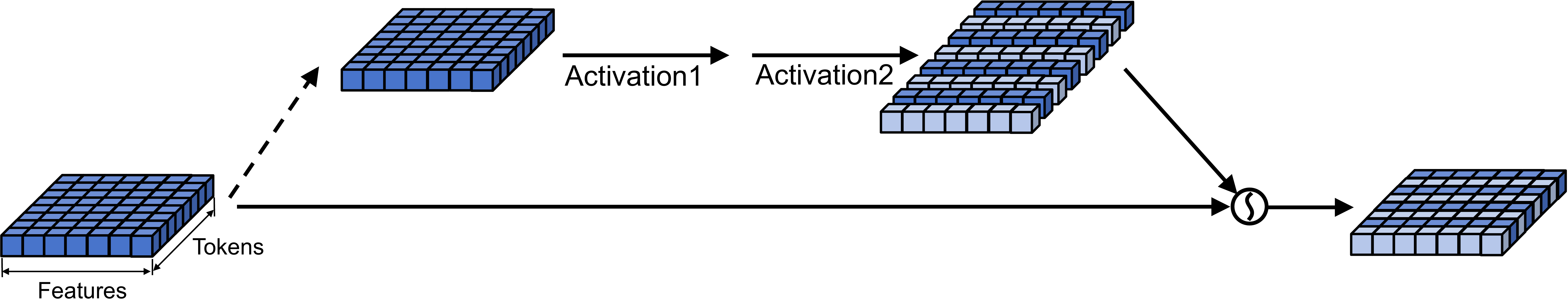}
        \caption{Traditional self-gate mechanism (consistent with the residual connection in Transformer).}
        \label{fig.sg1}
    \end{minipage}
    
    \vspace{8pt} % 专业排版建议间距
    
    \begin{minipage}[t]{0.9\linewidth}
        \centering
        \includegraphics[width=\linewidth]{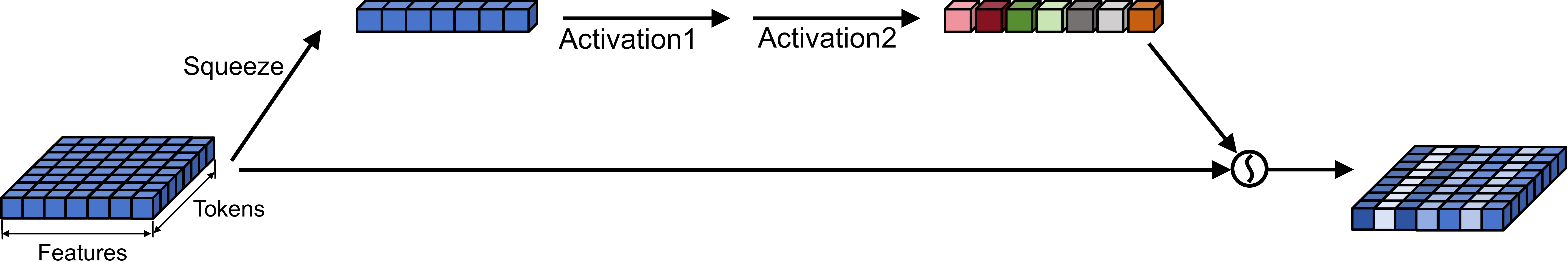}
        \caption{spatial squeeze self-gate mechanism.}
        \label{fig.sg2}
    \end{minipage}
    
    \label{fig.sg_group} % 整体分组标签
\end{figure}

The key to achieving cross-block sentence representation fusion lies in the calculation of adaptive weights, and the self-gate mechanism is the standard means to achieve adaptivity. Traditional self-gate methods obtain token-level activations as adaptive weights to guide the encoder to focus on important tokens in the sentence, as shown in Figure~\ref{fig.sg1}. However, these methods' activation of token representations and fusion mechanisms are essentially consistent with the feed-forward neural network and residual connection in Transformer, which undoubtedly leads to redundant information. 

Therefore, we advocate adopting the squeeze-excitation concept, which compresses the representation space to the feature-level representation and models dependencies at the feature level. As shown in Figure~\ref{fig.sg2}, using the activated compressed feature representation as adaptive weights can guide the focus on features meaningful to sentence representation, an angle of modeling not covered by Transformer. Undoubtedly, this low-information-redundancy self-gate mechanism is the first choice as adaptive fusion weights for sentence representation.

\subsubsection{SS Module}
Spatial Selection is a selective multiscale adaptive fusion mechanism with the input $\{\mathbf{u}^{(n)}\}_{n=0}^{N-1}$ and output $\mathbf{v}$. It consists of three stages: concatenation, squeeze-and-excitation, and selection. To capture the nonlinear transformations at the block-token level during the concatenation stage, we propose stacking the sentence representations of each block to build a 3D semantic space $\mathbf{U} \in \mathbb{R}^{N \times L \times D}$, as formulated in Equation~\ref{eq.fuse}.

\begin{equation}
    \mathbf{U}_{n,l}=[\mathbf{u}_l^{(0)};\cdots;\mathbf{u}_l^{(n)}]~\label{eq.fuse}
\end{equation}

% \begin{figure}[t]
%     \centering
%     \begin{subfigure}[b]{0.9\linewidth} % 添加对齐参数[b]
%         \includegraphics[width=\linewidth]{pictures/self_gate1.pdf}
%         \caption{The Traditional Self-Gate Mechanism (consistent with the residual connection in Transformer)}
%         \label{fig.sg1}
%     \end{subfigure}
    
%     \vspace{4pt} % 更合理的垂直间距
    
%     \begin{subfigure}[b]{0.9\linewidth}
%         \includegraphics[width=\linewidth]{pictures/self_gate2.pdf}
%         \caption{Our Spatial Squeeze based Self-Gate Mechanism}
%         \label{fig.sg2}
%     \end{subfigure}
    
%     \caption{Traditional and our Self-Gate mechanisms}
%     \label{fig.sg}
% \end{figure}

%接下来的squeeze-and-excitation阶段，我们采用嵌入在Spatial Selection中的Frequency Selection去进行空间squeeze。FS是对空间信息的压缩获得特征表示，$\mathbf{U}_{n,l}$和$\mathbf{v}_d$和分别为空间、特征表示符。

In the subsequent squeeze-and-excitation stage, we utilize Frequency Selection to perform spatial squeeze, as formulated in Equation~\ref{eq.fs}. Here, $\mathbf{f} \in \mathbb{R}^{D}$ denotes the squeezed feature representation, and we will discuss the specific methods of FS in detail in Section~\ref{sec.fs}.

\begin{equation}
    \mathbf{f}_d = {\rm FS}(\mathbf{U}_{n,l,d})\label{eq.fs}
\end{equation}

To capture the excited features and adaptively adjust their importance in the overall semantics, we employ a dimensional-reduction fully connected layer ($\mathbf{W}_1$) and dimensional-augmentation fully connected layers ($\mathbf{W}_2^{(n)}$) with the same number of Transformer blocks in the excitation stage. This allows us to obtain the excited vectors $\{\mathbf{e}^{(n)}\}_{n=0}^{N-1}$, where $\mathbf{e}^{(n)}\in \mathbb{R}^{ D}$, as formulated in Equation~\ref{eq.exc}. The surplus vector $\mathbf{s}=\max(0,\mathbf{f}\mathbf{W}_1)\in \mathbb{R}^{ \frac{D}{r}}$ forms a low-rank bottleneck. Here, $\delta$, $\sigma$ represent the Tanh, Sigmoid functions respectively, and $r$ acts as a reduction ratio that creates a bottleneck structure for parameter redundancy reduction.

\begin{equation}
    \mathbf{e}_d^{(n)} = \frac{1}{1+\exp(-{\max(0,\mathbf{f}_d\mathbf{W}_1)\mathbf{W}_2^{(n)}})}\label{eq.exc}
\end{equation}

%在selection阶段我们对激励向量$\{\mathbf{e}^{(n)}\}_{n=0}^{N-1}$进行softmax归一化（$\mathcal{F}_{\text{S-Norm}}$），归一化后的结果$\{\mathbf{e}^{(n)}\}_{n=0}^{N-1}$作为不同分支的权重，分别和各分支的表示$\{\mathbf{x}^{(n)}\}_{n=0}^{N-1}$元素级乘法并加和，得到各分支的加权求和的表示$\mathbf{u}=[\mathbf{u}_1,\mathbf{u}_2\cdots,\mathbf{u}_D] \in \mathbb{R}^{ D}$。

In the selection stage, we perform softmax normalization on the excitation vectors $\{\mathbf{e}^{(n)}\}_{n=0}^{N-1}$. The normalized results serve as weights for different branches. These weights are element-wise multiplied with the corresponding hidden states $\{\mathbf{u}^{(n)}\}_{n=0}^{N-1}$ and summed to obtain $\mathbf{v}_d$ as the final sentence representation.

\begin{equation}
\begin{aligned}
        \mathbf{v}_d = \frac{\sum_{n=0}^{N-1}\exp(\mathbf{e}^{(n)}_d)*\mathbf{u}^{(n)}_d}{\sum_{n=0}^{N-1}\exp(\mathbf{e}^{(n)}_d)}
\end{aligned}
\end{equation}     

\subsection{Frequency Selection}\label{sec.fs}

\subsubsection{DCT: Extension of GAP for Spatial Squeeze}
The above mentioned FS module nested within SS performs spatial squeeze to obtain feature representations, where $\bm {U} \in \mathbb{R}^{N \times L \times D}$ and $\mathbf{f} \in \mathbb{R}^D$. Inspired by channel attention, a classic approach is to use GAP to average the features across the $N$ and $L$ dimensions. However, this approach undoubtedly reduces the feature differences at different spatial levels, leading to inevitable semantic loss. We will demonstrate in this section that DCT is a better choice for spatial squeeze compared to GAP.

DCT is a widely used data compression method in signal processing, particularly for digital images and videos. First, let us review the general form of a DCT as formulated in Equation~\ref{eq.dct_gen}. Here, $\mathbf{\alpha}$ represents the corresponding DCT frequency basis functions. It is worth noting that, for the sake of simplicity, the normalization constant factor of DCT has been omitted.

\vspace{-8pt}
\begin{small}
\begin{equation}
\begin{aligned}
&\text{DCT}(\mathbf{X}_{i,j})=\sum_{i=0}^{N-1}\sum_{j=0}^{L-1}\mathbf{X}_{i,j}\mathbf{\alpha}^{i,j}_{n,l},\\
&\mathbf{\alpha}^{i,j}_{n,l}=\cos(\frac{\pi n}{N}(i+\frac{1}{2}))\cos(\frac{\pi l}{L}(j+\frac{1}{2}))\\
& \text{Inverse Transform:}\quad \mathbf{X}_{i,j}={\rm IDCT}\sum_{n=0}^{N-1}\sum_{l=0}^{L-1}\mathbf{x}_{n,l}\mathbf{\alpha}^{i,j}_{n,l}
\label{eq.dct_gen}    
\end{aligned}
\end{equation}
\end{small}
To relate GAP to DCT, we extend the representation of GAP as formulated in Equation~\ref{eq.gap}. 

% 
% \begin{equation}
% \begin{aligned}
% &\mathbf{x}_{\textcolor{blue}{0},\textcolor{blue}{0}}=\sum_{i=0}^{N-1}\sum_{j=0}^{L-1}\mathbf{X}_{i,j}\mathbf{\alpha}^{i,j}_{\textcolor{blue}{0},\textcolor{blue}{0}}\\&=\sum_{i=0}^{N-1}\sum_{j=0}^{L-1}\mathbf{X}_{i,j}\cos(\frac{\pi*\textcolor{blue}{0}}{N}(i+\frac{1}{2}))\cos(\frac{\pi*\textcolor{blue}{0}}{L}(j+\frac{1}{2}))\\&=\sum_{i=0}^{N-1}\sum_{j=0}^{L-1}\mathbf{X}_{i,j}=NL*\mathcal{F}_{\rm GAP}(\mathbf{X}_{i,j})\label{eq.gap}    
% \end{aligned}    
% \end{equation}

\vspace{-8pt}
\begin{small}
\begin{equation}
\begin{aligned}
&{\rm GAP}(\mathbf{X}_{i,j})=\!\!\frac{1}{NL}\sum_{i=0}^{N-1}\sum_{j=0}^{L-1}\mathbf{X}_{i,j}\\&=\!\!\frac{1}{NL}\sum_{i=0}^{N-1}\sum_{j=0}^{L-1}\mathbf{X}_{i,j}\cos(\frac{\pi*\textcolor{blue}{0}}{N}(i+\frac{1}{2}))\cos(\frac{\pi*\textcolor{blue}{0}}{L}(j+\frac{1}{2}))\\&=\!\!\frac{1}{NL}\sum_{i=0}^{N-1}\sum_{j=0}^{L-1}\mathbf{X}_{i,j}\mathbf{\alpha}^{i,j}_{\textcolor{blue}{0},\textcolor{blue}{0}}=\!\!\frac{1}{NL}\mathbf{x}_{\textcolor{blue}{0},\textcolor{blue}{0}}\label{eq.gap}    
\end{aligned}
\end{equation}
\end{small}
$\mathbf{x}_{\textcolor{blue}{0},\textcolor{blue}{0}}$ represents the lowest frequency component of DCT and is directly proportional to GAP. After establishing this point, we can rewrite the inverse transformation of DCT as formulated in Equation~\ref{eq.dct-1}, which naturally decomposes the semantic spatial information into a combination of different frequency components. 

\vspace{-8pt}
\begin{small}
\begin{equation}
\begin{aligned}
&\mathbf{X}_{i,j}=\sum_{n=0}^{N-1}\sum_{l=0}^{L-1}\mathbf{x}_{n,l}\mathbf{\alpha}^{i,j}_{n,l}\\&=
\mathbf{x}_{0,0}\mathbf{\alpha}^{i,j}_{0,0}+\mathbf{x}_{0,1}\mathbf{\alpha}^{i,j}_{0,1}+\cdots+\mathbf{x}_{N-1,L-1}\mathbf{\alpha}^{i,j}_{N-1,L-1}\\&=
NL{\rm GAP}(\mathbf{X}_{i,j})\mathbf{\alpha}^{i,j}_{0,0}+\cdots+\mathbf{x}_{N-1,L-1}\mathbf{\alpha}^{i,j}_{N-1,L-1}
\label{eq.dct-1}
\end{aligned}
\end{equation}
\end{small}

From the perspective of feature descriptor, Equation~\ref{eq.dct---1} shows that the representation of each feature under the GAP operation only utilizes the lowest frequency component (\Checkmark) of the DCT, while discarding additional higher frequency components (\XSolidBrush). This implies that compared to GAP, DCT can explore a broader range of high-frequency components.

\begin{small}
\begin{equation}
\begin{aligned}
&\mathbf{X}_{:;:;d}\!\!=
\begin{bmatrix}
\underbrace{\mathcal{P}\mathbf{\alpha}^{0,0}_{0,0}}_{\text{\Checkmark}}+\underbrace{\mathcal{D}^{0,0}}_{\text{\XSolidBrush}} & \!\!\!\cdots & \!\!\!\underbrace{\mathcal{P}\mathbf{\alpha}^{0,L-1}_{0,0}}_{\text{\Checkmark}}+\underbrace{\mathcal{D}^{0,L-1}}_{\text{\XSolidBrush}} \\
\vdots & \!\!\!\ddots & \!\!\!\vdots \\
\underbrace{\mathcal{P}\mathbf{\alpha}^{N-1,0}_{0,0}}_{\text{\Checkmark}}+\underbrace{\mathcal{D}^{N-1,0}}_{\text{\XSolidBrush}} & \!\!\!\cdots & \!\!\!\underbrace{\mathcal{P}\mathbf{\alpha}^{N-1,L-1}_{0,0}}_{\text{\Checkmark}}+\underbrace{\mathcal{D}^{N-1,L-1}}_{\text{\XSolidBrush}} \\
\end{bmatrix},\\
&\mathcal{P} = NL{\rm GAP}(\mathbf{X}_{i,j}),\\&\mathcal{D}^{i,j}=\mathbf{x}_{0,1}\mathbf{\alpha}^{i,j}_{0,1}+\cdots+\mathbf{x}_{N-1,L-1}\mathbf{\alpha}^{i,j}_{N-1,L-1}\label{eq.dct---1}
\end{aligned}
\end{equation}
\end{small}

\subsubsection{FS Module}

%Frequency Selection ($\mathcal{F}_{\rm FS}:\mathbf{U}\rightarrow\mathbf{f}$)是一种多光谱选择性的空间压缩，通过定制离散余弦变化（DCT）进行空间压缩获取语义特征表示，并为每一个特征选择合适的DCT频率分量。对于输入的$\mathbf{U}$,我们先按照特征维度将其split成$m$个部分，如Equationref{eq.split}。其中$\mathbf{U}^{[m]}_{n,l}\in\mathbb{R}^{L \times \frac{D}{M}}$。

Frequency Selection is a multi-spectral with the input $\mathbf{U}$ and the output $\mathbf{f}$. It selects appropriate DCT frequency basis functions for each feature. For the input $\mathbf{U}$, we first split it into $m$ parts along the feature dimension, as formulated in Equation~\ref{eq.split}. Here, $\mathbf{U}^{[m]}_{n,l}\in\mathbb{R}^{N\times L \times \frac{D}{M}}$.

\begin{equation}
    \mathbf{U}_{n,l}\xrightarrow{\text{Split}}<\mathbf{U}^{[0]}_{n,l};\cdots;\mathbf{U}^{[m]}_{n,l}>\label{eq.split}
\end{equation}

%对于每个部分$\mathbf{U}^{[m]}$,采用DCT进行空间压缩，为每个部分选择一个相应的二维DCT频率分量，如Equation~\ref{eq.dct1}所示，其中$a_m,b_m$为$\mathbf{U}^{[m]}$对应的频率分量二维指数。相关研究已经表明cnn更倾向于低频信息。所以我们在FS module中选取最低频的$m$个频率分量。我们在实验部分分别讨论了$m$取1（GAP），2，4，8，16的情况，具体的选择机制如图~\ref{fig.lf}所示。其中，$\mathbf{v}\in\mathbb{R}^{ \frac{D}{M}}$是压缩后的句嵌入特征表示。

\begin{figure}[t]
\centering
\includegraphics[width=1\linewidth]{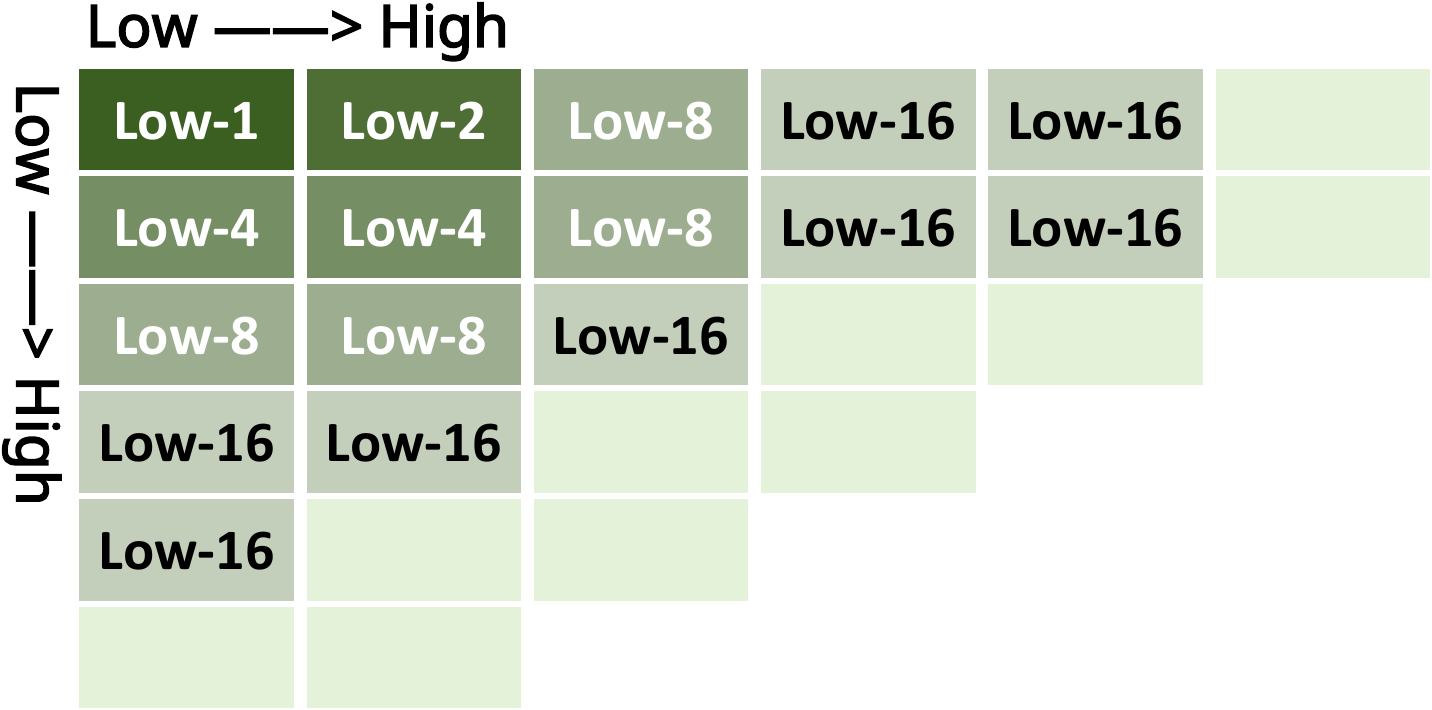}
\caption{Selection for $m$ low frequency basis functions
}\label{fig.lf}
\end{figure}

For each part $\mathbf{U}^{[m]}$, we perform spatial squeeze using selected 2D DCT basis functions, as formulated in Equation~\ref{eq.dct1}, where $a_m$ and $b_m$ are the two-dimensional indices of the frequency basis functions corresponding to $\mathbf{U}^{[m]}$. Previous studies~\cite{hu2018squeeze,xu2020learning} have shown that neural networks tend to be biased towards low-frequency information. Therefore, in the FS module, we select the lowest $m$ frequency basis functions. In the experimental section,  we explore different values of $m$ from \{1 (GAP), 2, 4, 8, 16\}. The specific selection mechanism is illustrated in Figure~\ref{fig.lf}. Here, $\mathbf{f}^{[m]}\in\mathbb{R}^{\frac{D}{M}}$ denotes the squeezed feature representation of $m^{\rm th}$ part.

\begin{equation}
\begin{aligned}
    \mathbf{f}^{[m]}_{d'} &= {\rm DCT}^{a_m,b_m}(\mathbf{U}^{[m]})\\&=\sum_{n=0}^{N-1}\sum_{l=0}^{L-1}\mathbf{U}^{[m]}_{n,l,d'}\mathbf{\alpha}_{n,l}^{a_m,b_m}\label{eq.dct1}    
\end{aligned}
\end{equation}

%最后整体的嵌入表示将$\mathbf{v}^{[m]}$按照维度进行串联，如Equation~\ref{eq.f}所示。其中<;>是按照特征维度串联操作。

\begin{table*}[t]
\centering
\footnotesize
\renewcommand{\arraystretch}{1.1}
\setlength{\tabcolsep}{2.5 pt}
\begin{tabular}{lllcccccccc}
\hline\hline
\multirow{2}{*}{\textbf{Backbone}} & \multirow{2}{*}{\textbf{Block}} & \multirow{2}{*}{\textbf{Downstream}} & \multirow{2}{*}{$\Delta$\textbf{Param.}} & \multirow{2}{*}{$\Delta$\textbf{Latency}} & \multicolumn{3}{c}{\textbf{SimCSE}} & \multicolumn{3}{c}{\textbf{PromptCSE}} \\ \cline{6-11}
& & & & & \multicolumn{1}{c}{\textbf{STS-B}} & \multicolumn{1}{c}{\textbf{SICK-R}} & \textbf{STS-Avg.} & \multicolumn{1}{c}{\textbf{STS-B}} & \multicolumn{1}{c}{\textbf{SICK-R}} & \textbf{STS-Avg.} \\ \hline
\multirow{8}{*}{BERT} & \multirow{2}{*}{Last1} & --Base & - & - & \multicolumn{1}{c}{76.88} & \multicolumn{1}{c}{72.19} & 76.21 & \multicolumn{1}{c}{81.70} & \multicolumn{1}{c}{69.57} & 78.11 \\
& & --S\textsuperscript{2}Sent(1D SS) & 0.08\% & 0.20\% & \multicolumn{1}{c}{{\ul{77.82}}} & \multicolumn{1}{c}{{\ul{71.98}}} & 77.18 & \multicolumn{1}{c}{{\ul{82.34}}} & \multicolumn{1}{c}{69.78} & {\ul{79.85}} \\ \cline{2-11}
& \multirow{3}{*}{Last3} & --Avg. & 0.00\% & 0.00\% & \multicolumn{1}{c}{77.25} & \multicolumn{1}{c}{72.05} & 76.48 & \multicolumn{1}{c}{82.06} & \multicolumn{1}{c}{69.89} & 79.22 \\
& & --Avg.--S\textsuperscript{2}Sent(1D SS) & 0.08\% & 0.20\% & \multicolumn{1}{c}{{\ul{77.79}}} & \multicolumn{1}{c}{{\ul{72.05}}} & {\ul{77.02}} & \multicolumn{1}{c}{{\ul{82.62}}} & \multicolumn{1}{c}{69.62} & {\ul{79.71}} \\
& & --Stack--S\textsuperscript{2}Sent(2D SS) & 0.17\% & 0.20\% & \multicolumn{1}{c}{\textbf{79.54}} & \multicolumn{1}{c}{\textbf{73.18}} & {\ul{\textbf{78.50}}} & \multicolumn{1}{c}{{\ul{83.41}}} & \multicolumn{1}{c}{{\ul{\textbf{71.15}}}} & {\ul{\textbf{80.98}}} \\ \cline{2-11}
& \multirow{3}{*}{Last6} & --Avg. & 0.00\% & 0.00\% & \multicolumn{1}{c}{{\ul{76.73}}} & \multicolumn{1}{c}{72.21} & 76.17 & \multicolumn{1}{c}{{\ul{81.81}}} & \multicolumn{1}{c}{{\ul{69.92}}} & 79.50 \\
& & --Avg.--S\textsuperscript{2}Sent(1D SS) & 0.08\% & 0.20\% & \multicolumn{1}{c}{{\ul{77.87}}} & \multicolumn{1}{c}{{\ul{71.75}}} & {\ul{77.06}} & \multicolumn{1}{c}{{\ul{82.19}}} & \multicolumn{1}{c}{{\ul{69.82}}} & {\ul{79.90}} \\
& & --Stack--S\textsuperscript{2}Sent(2D SS) & 0.34\% & 0.20\% & \multicolumn{1}{c}{{\ul{79.23}}} & \multicolumn{1}{c}{{\ul{73.03}}} & {\ul{78.14}} & \multicolumn{1}{c}{{\ul{\textbf{83.58}}}} & \multicolumn{1}{c}{{\ul{70.97}}} & {\ul{80.81}} \\ \hline\hline
\multirow{8}{*}{RoBERTa} & \multirow{2}{*}{Last1} & --Base & - & - & \multicolumn{1}{c}{80.32} & \multicolumn{1}{c}{68.51} & 76.46 & \multicolumn{1}{c}{81.81} & \multicolumn{1}{c}{69.74} & 79.02 \\
& & --S\textsuperscript{2}Sent(1D SS) & 0.07\% & 0.19\% & \multicolumn{1}{c}{{\ul{81.68}}} & \multicolumn{1}{c}{{\ul{69.25}}} & {\ul{78.19}} & \multicolumn{1}{c}{{\ul{82.17}}} & \multicolumn{1}{c}{{\ul{70.34}}} & {\ul{79.80}} \\ \cline{2-11}
& \multirow{3}{*}{Last3} & --Avg. & 0.00\% & 0.00\% & \multicolumn{1}{c}{{\ul{80.73}}} & \multicolumn{1}{c}{68.47} & 76.61 & \multicolumn{1}{c}{81.97} & \multicolumn{1}{c}{{\ul{69.89}}} & {\ul{79.61}} \\
& & --Avg.--S\textsuperscript{2}Sent(1D SS) & 0.07\% & 0.19\% & \multicolumn{1}{c}{{\ul{81.97}}} & \multicolumn{1}{c}{{\ul{69.11}}} & {\ul{78.35}} & \multicolumn{1}{c}{81.98} & \multicolumn{1}{c}{69.82} & {\ul{80.36}} \\
& & --Stack--S\textsuperscript{2}Sent(2D SS) & 0.15\% & 0.19\% & \multicolumn{1}{c}{{\ul{\textbf{82.57}}}} & \multicolumn{1}{c}{{\ul{\textbf{70.95}}}} & {\ul{78.32}} & \multicolumn{1}{c}{{\ul{\textbf{83.91}}}} & \multicolumn{1}{c}{{\ul{\textbf{71.21}}}} & {\ul{81.18}} \\ \cline{2-11}
& \multirow{3}{*}{Last6} & --Avg. & 0.00\% & 0.00\% & \multicolumn{1}{c}{80.13} & \multicolumn{1}{c}{68.60} & 76.53 & \multicolumn{1}{c}{81.77} & \multicolumn{1}{c}{{\ul{69.98}}} & {\ul{79.50}} \\
& & --Avg.--S\textsuperscript{2}Sent(1D SS) & 0.07\% & 0.19\% & \multicolumn{1}{c}{{\ul{81.81}}} & \multicolumn{1}{c}{{\ul{69.23}}} & {\ul{78.26}} & \multicolumn{1}{c}{82.08} & \multicolumn{1}{c}{70.34} & 79.98 \\
& & --Stack--S\textsuperscript{2}Sent(2D SS) & 0.30\% & 0.19\% & \multicolumn{1}{c}{{\ul{82.41}}} & \multicolumn{1}{c}{{\ul{70.89}}} & {\ul{\textbf{78.48}}} & \multicolumn{1}{c}{{\ul{83.67}}} & \multicolumn{1}{c}{{\ul{71.57}}} & {\ul{\textbf{81.25}}} \\ \hline\hline
\end{tabular}
\caption{Study for SS, evaluation results on STS tasks}
\label{tab.main}
\end{table*}

Finally, we concatenate the $m$ part feature representations $\mathbf{f}_{d'}^{[m]}$ along the dimensions to obtain the final squeezed feature representation, as formulated in Equation~\ref{eq.f}.

\begin{equation}
    \mathbf{f}_d = <\mathbf{f}^{[0]}_{d'};\cdots;\mathbf{f}^{[m]}_{d'}>\label{eq.f}
\end{equation}

\section{Experimental Setup}

\subsection{Implementation}

%我们选择了Backbone BERT~\cite{devlin2019bert}和RoBERTa~\cite{liu2019roberta}的base版本作为我们的Backbone。我们采用无监督的对比学习训练句子embedding，具体来说，我们采用了SimCSE和PromptCSE作为我们的基线。我们使用所有token embedding的平均作为句子的1D表示。在训练时，我们对SimCSE设置学习率为3e-5，对PromptCSE设置学习率为1e-5。批量大小为64，训练温度$\tau$为0.05。对于BERT，我们将组大小的数量设置为384，对于RoBERTabase和我们将组大小的数量设置为256。我们将所有模型的正数设为3。我们在STS-B的dev上每125个step对模型进行一次评估，并保持对测试集进行评估的最佳检查点。我们在6个V100的GPU上进行了实验。

%We chose BERT~\cite{devlin2019bert} and RoBERTa~\cite{liu2019roberta} as our backbone. We used two  unsupervised contrastive learning methods SimCSE~\cite{gao2021simcse} and PromptCSE~\cite{jiang2022promptbert} to train sentence embeddings. We used the average of all token embeddings as the 1D sentence representation. During training, we set the learning rate to 3e-5 for SimCSE and 1e-5 for PromptCSE, the batch size to 64, and the training temperature $\tau$ to 0.05. For BERT, we set the number of groups to 384, and for RoBERTa, we set the number of groups to 256. We set the positive number for all models to 3. We evaluated the model every 125 steps on the dev set of STS-B and kept the best checkpoint for evaluating on the test set. We conducted experiments on seven V100 GPUs.

We used BERT~\cite{devlin2019bert} and RoBERTa~\cite{liu2019roberta} as the backbone models. For training sentence embeddings, we employed two unsupervised contrastive learning methods: SimCSE~\cite{gao2021simcse} and PromptCSE~\cite{jiang2022promptbert}. We represented sentences as the average of all token embeddings. During training, the learning rate was set to \(3 \times 10^{-5}\) for SimCSE and \(1 \times 10^{-5}\) for PromptCSE. We used a batch size of 64 and set the training temperature \(\tau\) to 0.05. For BERT, we used 384 groups, while for RoBERTa, we used 256 groups. The positive number was set to 3 for all models. The model was evaluated every 125 steps on the STS-B dev set, and the best checkpoint was used for testing. Experiments were conducted on seven V100 GPUs.

% \subsection{Competitive Methods}

% %我们其他具有竞争力的方法进行了对比试验：SBERT~\cite{reimers2019sentence}、Flow~\cite{li2020sentence}、Whitening~\cite{su2021whitening}、IS~\cite{zhang2020unsupervised}、CT~\cite{carlsson2021semantic}、ConSERT~\cite{yan2021consert}、SimCSE~\cite{gao2021simcse}、MixCSE~\cite{zhang2022unsupervised}、ArcCSE~\cite{zhang2021zero}、DCLR~\cite{zhou2022debiased}、Denosent~\cite{wang2024denosent}.其中所有算法都是基于无监督条件下进行训练的。

%  We conducted comparative experiments with other competitive baselines: Sentence BERT~\cite{reimers2019sentence}, BERT-Flow~\cite{li2020sentence}, BERT-Whitening~\cite{su2021whitening}, IS-BERT~\cite{zhang2020unsupervised}, CT-BERT~\cite{carlsson2021semantic}, ConSERT-BERT~\cite{yan2021consert}, SimCSE-BERT~\cite{gao2021simcse}, MixCSE-BERT~\cite{zhang2022unsupervised}, ArcCSE-BERT~\cite{zhang2021zero}, DCLR-BERT~\cite{zhou2022debiased}, WhitenedCSE-BERT~\cite{zhuo2023whitenedcse}, Denosent-BERT~\cite{wang2024denosent}. All of these algorithms were trained under unsupervised conditions.

\subsection{Benchmarks}

%我们将在Semantic Textual Similarity（STS）任务上评估我们的方法,我们使用SentEval~\cite{conneau2018senteval}来评估所有的任务。语义文本相似度（STS）任务由7个任务组成：STS2012-2016~\cite{agirre2012semeval,agirre2013sem,agirre2014semeval,agirre2015semeval,agirre2016semeval}, STS Benchmark~\cite{cer2017semeval}和SICK Relatedness~\cite{marelli2014semeval}。这些数据集中的每个样本都有两个句子和一个注释的0-5相似度评分。

We evaluated our method on the Semantic Textual Similarity (STS) task using SentEval~\cite{conneau2018senteval}, which provides a comprehensive evaluation framework for multiple tasks.
The STS tasks consists of: STS 2012-2016~\cite{agirre2012semeval,agirre2013sem,agirre2014semeval,agirre2015semeval,agirre2016semeval}, STS Benchmark~\cite{cer2017semeval}, and SICK Relatedness~\cite{marelli2014semeval}. 

\section{Experimental Results}

\subsection{Main Results}

%在这一部分，我们用两个backbone BERT、RoBERTa和两个训练baseline SimCSE、PromptCSE构成了4组基线。我们通过控制变量来讨论S\textsuperscript{2}Sent的两个子模块SS和FS对于这4组句子表示基线的影响。

In this section, we utilized two backbones, BERT and RoBERTa, along with two sentence embedding training methods, SimCSE and PromptCSE, to establish four baseline models. By controlling variables, we examined the impact of the two selections of S\textsuperscript{2}Sent, SS and FS, on these baseline models for sentence representation.

\subsubsection{Effect of Spatial Selection}
%S\textsuperscript{2}Sent的本质即是嵌入了频率选择的空间选择，我们首先控制FS不变，探究SS结构为基线方法带来的影响。Table~\ref{tab.main}报告了在STS-Benchmark、SICK-Relatedness、以及7个benchmark的斯皮尔曼系数平均值,值越接近1，两个统计变量越接近正相关。在每组实验中，--Reported表示的是官方的实现结果，--Base作为我们实现的仅用Last1作为表示的基线结果，我们讨论了采用不同的block（Last1，Last3，Last6）的表示以及不同的下游句子表示策略对基线结果的影响。1、首先我们可以看到，对于Last3、Last6策略，简单地用平均作为表示无法显著提高仅用Last1作为表示的基线结果。然而，在平均后的表示后引入S\textsuperscript{2}Sent(1D SS)进行表示优化,其表现在各组实验上都产生了显著的提升，这无疑体现了S\textsuperscript{2}Sent中SS的有效性；进一步地，S\textsuperscript{2}Sent的标准实现(2D),即将不同block的表示Stacked Concatenate，在所有组别中均取得了最好的结果。这说明Stack Concatenate各表示用3D张量进行下游建模能够更加充分得保留不同Transformer中层次化的语义信息。

%We first controlled FS and explore the impact of SS to 4 baselines. Table~\ref{tab.main} reports the Spearman correlation coefficients for STS-Benchmark, SICK-Relatedness, and the average value of seven benchmarks. The closer Spearman correlation coefficient is to 1, the closer the two statistical variables are positively correlated. In each set of experiments, --Reported denotes the reported implementation results in previous papers, while --Base denotes our recalculated baseline results using Last1 block only as the representation.
We first controlled the FS module and explored the impact of SS on the four baseline models. Table~\ref{tab.main} reports the Spearman correlation coefficients for STS-Benchmark, SICK-Relatedness, and the average across seven benchmarks, while also presenting the additional parameters and inference latency introduced. A higher Spearman correlation coefficient indicates a stronger positive correlation between the two statistical variables. In each set of experiments, --Base refers to our recalculated baseline results using only the Last1 block for representation.
Table~\ref{tab.main} also presents the effects of using different block representations (Last1, Last3, Last6) and various downstream sentence representation strategies on the baseline results. Our experimental results are averaged over 7 random seeds. The \underline{value} indicates a significant improvement over the base (based on t-tests), and the \textbf{value} represents the best result in the group.

%It's obviously that for Last3 and Last6 strategies, simply using the average as a representation does not significantly improve the base set using Last1 as representation; However, after introducing S\textsuperscript{2}Sent(1D SS) for representation optimization after averaging, its performance has significantly improved in all sets of experiments, which undoubtedly reflects the effectiveness of SS in S\textsuperscript{2}Sent;
It is evident that for the Last3 and Last6 strategies, simply averaging the representations does not significantly improve over the base set using Last1. However, introducing S\textsuperscript{2}Sent(1D SS) for representation optimization after averaging leads to significant improvements across all experimental sets, demonstrating the effectiveness of SS in S\textsuperscript{2}Sent.
%Further, the standard implementation of S\textsuperscript{2}Sent(2D SS), which stacks concatenate representations of different blocks, achieves the best results in all groups. This indicates that using 3D tensors for downstream modeling with stack concatenate representations can more fully preserve the hierarchical semantic information in different Transformers blocks.
Furthermore, the standard implementation of S\textsuperscript{2}Sent(2D SS), which stacks and concatenates representations from different blocks, achieves the best results in all groups. This indicates that using 3D tensors for downstream modeling with stacked and concatenated representations better preserves the hierarchical semantic information across different Transformer blocks.
Based on the reported additional overhead of S\textsuperscript{2}Sent, we can observe that the introduced parameters amount to less than 0.5\% of the backbone, which is superior to the bottleneck layer set in SS. Similarly, the inference latency introduced is extremely low, as it does not involve high-cost attention mechanisms like cross-attention, and the inference of multiple branches can be processed in parallel. Overall, S\textsuperscript{2}Sent is undoubtedly a lightweight module.

\subsubsection{Effect of Frequency Selection}

\begin{figure}[t]
\centering
\includegraphics[width=1\linewidth]{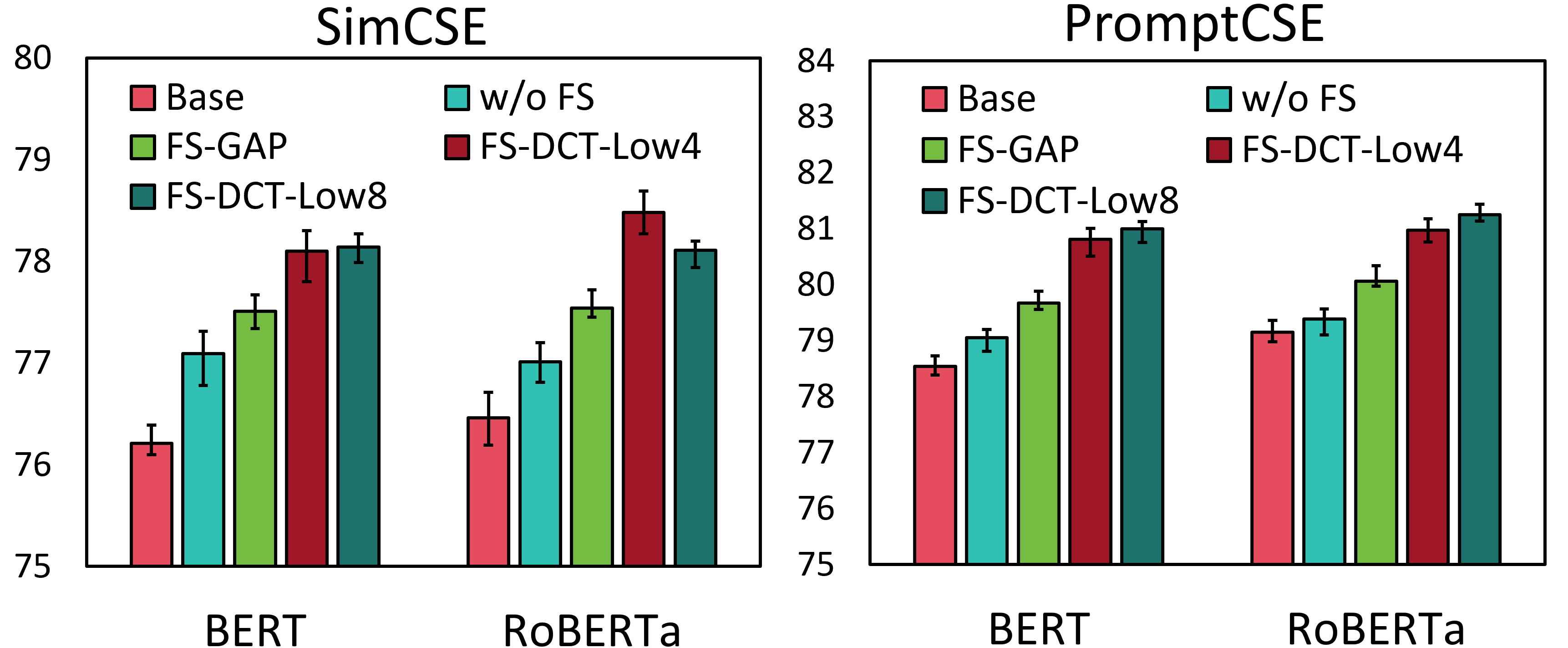}
\caption{Study for FS, evaluation results on STS tasks}
\label{fig.fs_ablation}
\end{figure}

%进一步地，我们控制SS中的各个条件均为最优的设置，在这一部分讨论FS的作用。我们讨论了1、没有FS进行空间squeeze、2、用GAP进行sqeeze，3、用Low-4和4、Low-8的DCT基函数进行squeeze这几种情况的影响。Figure~\ref{fig.fs_ablation}展示了各个情况下7个STS tasks的平均值，可以得到以下结论。1、当不采用FS进行空间squeeze，直接采用全连接层进行激活，S\textsuperscript{2}Sent对backbone的句子表征能力有一定的提升，这说明了SS进行多尺度自适应融合机制的有效性。2、当引入了FS进行空间压缩后，backbone的表示能力进一步地提升，这是因为激活空间压缩后地特征表示避免了直接激活token表示造成的与Transformer结构的信息冗余，同时建模了各个embedding特征间的依赖。3、采用DCT进行空间squeeze相较于GAP有着更加优异的表现。采用4个和8个低频的基函数在2个backbone和2个对比学习的组合中均取得了最好的表现。这说明了作为GAP的拓展，DCT有更好的特征通道标量化的潜力。

%Furthermore, we controlled all the conditions in SS to be optimally set, and in this section, we discuss the role of FS. Figure~\ref{fig.fs_ablation} discusses the impact of the following cases: (1) no FS for spatial squeeze, (2) using GAP for squeeze, (3) using Low-4, (4) Low-8 DCT basis functions for squeeze. Figure~\ref{fig.fs_ablation} shows the averages of the seven STS tasks under each case. 

Additionally, we controlled all conditions in SS to be optimally set, and in this section, we discuss the role of FS. Figure~\ref{fig.fs_ablation} illustrates the impact of the following cases: (1) no FS for spatial squeeze, (2) using GAP for squeeze, (3) using Low-4 DCT basis functions, and (4) using Low-8 DCT basis functions for squeeze. Figure~\ref{fig.fs_ablation} shows the averages across the seven STS tasks for each case.

%When FS is not used for spatial squeezing and activation is directly performed using fully connected layers, S\textsuperscript{2}Sent shows some improvement in the sentence representation capability of the backbone. This indicates the effectiveness of the multi-scale adaptive fusion mechanism.
%When FS is introduced for spatial squeeze, the representation capability of the backbone is further enhanced. This is because the feature representation by activating the spatial squeezed features models the dependencies between different embedding features.
%DCT-based spatial squeezing outperforms GAP. The use of 4 and 8 low-frequency basis functions achieves the best performance in combinations of two backbone models and two contrastive learning methods. This demonstrates that DCT has better potential for feature channel scalarization as an extension of GAP.

%hui liu: 时态是一个问题。对于实验设计的部分，是否都要用过去时？
When FS was not used for spatial squeezing and activation was directly performed using fully connected layers, S\textsuperscript{2}Sent showed some improvement in the sentence representation capability of the backbone. This indicates the effectiveness of the multi-scale adaptive fusion mechanism.
Introducing FS for spatial squeeze further enhanced the backbone's representation capability. This is because the feature representation, activated by the spatially squeezed features, models the dependencies between different embedding features.
DCT-based spatial squeezing outperforms GAP. Using 4 and 8 low-frequency basis functions achieved the best performance across combinations of two backbone models and two contrastive learning methods. This demonstrates that DCT offers better potential for spatial squeeze as an extension of GAP.
\subsection{Hyperparameter}

%Table~\ref{tab.main}和Figure~\ref{fig.fs_ablation}中初步展示了$m$为Last3、Last6的情况和$n$为Low4、Low8的情况，在这一部分我们进行更加全面的实验进行分析。图~\ref{fig.last_low}报告了选择的Transfomrer block的数量$n$和选择的低频DCT基函数数量对引入S\textsuperscript{2}Sent的BERT、RoBERTa的性能的影响。针对于超参数$n$，我们分别进行了取Transformer的$n=$last1，3，6，9，12 block的表示进行5组实验，每组实验根据基函数数量选择$m=$ low1，2，4，6，8获取5个样本点绘制箱线图，对于超参数$m$同理。根据图~\ref{fig.last_low}，我们知道当$n=$ last1和$m$=low1即是我们最常使用的Transformer last hidden作为表示以及GAP空间压缩。可以发现当$n$和$m$都大于1时，模型的表现都有显著提升且相互之间差异不大。这无疑是一个好消息，因为这在凸显我们方法的显著有效性的同时减小了对超参数选择的要求。

\subsubsection{Effect of $m$ \& $n$}
%Table~\ref{tab.main} and Figure~\ref{fig.fs_ablation} provide preliminary discussions of the cases where $m$ is Last3 and Last6, and $n$ is Low4 and Low8. In this section, we conduct more comprehensive experiments for in-depth analysis. Figure~\ref{fig.last_low} illustrates the impact of the selected number of Transformer blocks $n$, and the chosen number of low-frequency DCT basis functions $m$, on the performance of 4 baselines. For the hyperparameter $n$, we conducted five sets of experiments using representations from the last1, 3, 6, 9, 12 Transformer blocks, respectively. For each set of experiments, we selected $m$ from low1, 2, 4, 8, 16 to obtain five data points, and then plotted the box plots. The same procedure was applied to the hyperparameter $m$. 

%Figure~\ref{fig.last_low} illustrates that when $n$ is set to last1 and $m$ is set to low1, which corresponds to using the last hidden state of the Transformer as the representation and applying GAP for spatial squeeze, the model performs the worst in all cases. When $m$ and $n$ are set above 1, the model's performance significantly improves with little variation between them. This is undoubtedly good news as it highlights the effectiveness of our approach while reducing the reliance on hyperparameter selection.

\begin{figure}[t]
\centering
\includegraphics[width=1\linewidth]{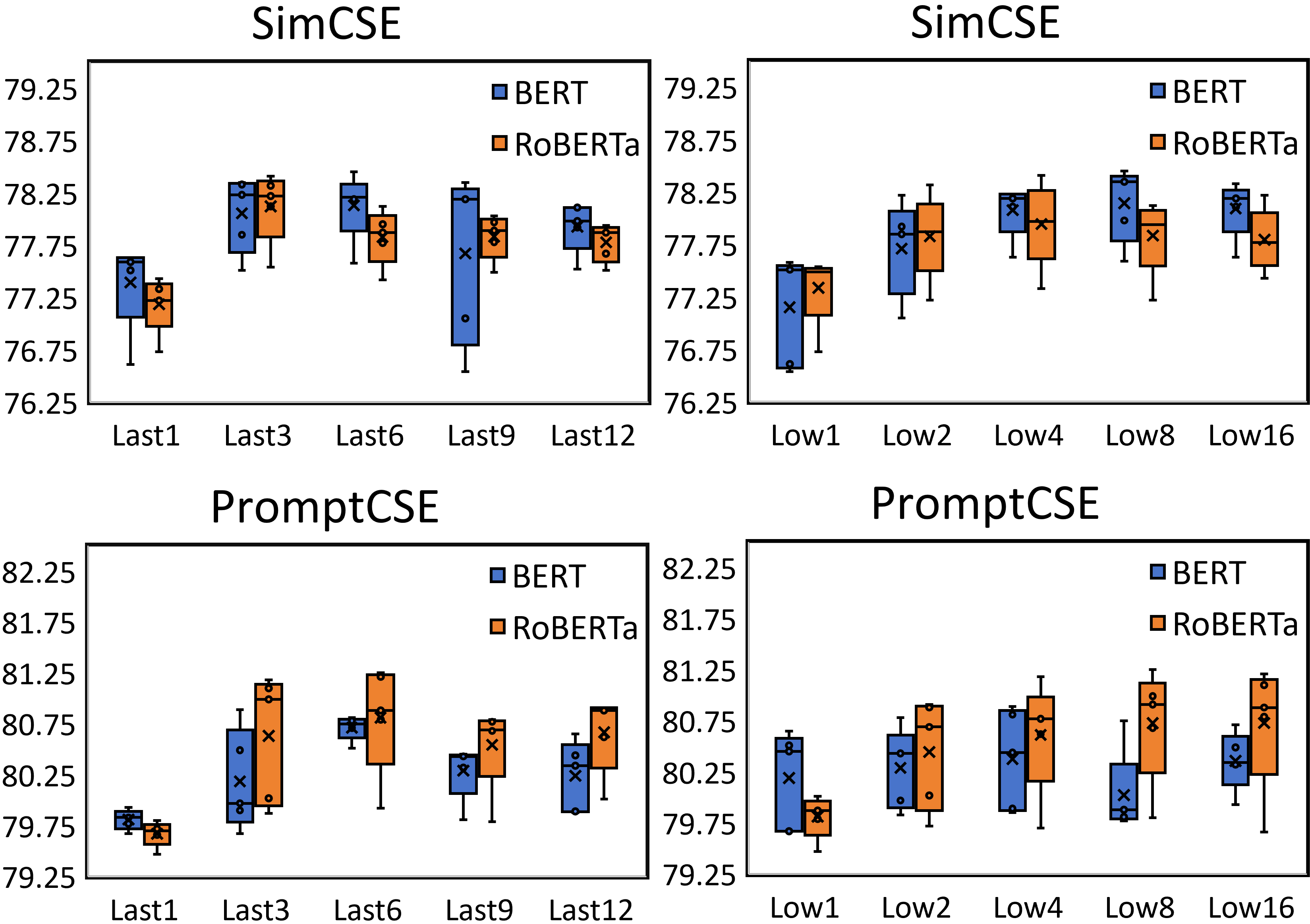}
\caption{The impact of $n$ (left) and $m$ (right).}
\label{fig.last_low}
\end{figure}

Table~\ref{tab.main} and Figure~\ref{fig.fs_ablation} offer preliminary insights into cases where $m$ is Last3 and Last6, and $n$ is Low4 and Low8. In this section, we conducted more comprehensive experiments for in-depth analysis. Figure~\ref{fig.last_low} illustrates the impact of the selected number of Transformer blocks ($n$) and the chosen number of low-frequency DCT basis functions ($m$) on the performance of four baselines. For the hyperparameter $n$, we conducted five sets of experiments using representations from the last1, 3, 6, 9, and 12 Transformer blocks, respectively. For each set, we selected $m$ from low1, 2, 4, 8, and 16 to obtain five data points, which were then plotted as box plots. The same procedure was applied to the hyperparameter $m$.

Figure~\ref{fig.last_low} shows that when $n$ was set to last1 and $m$ was set to low1, which corresponds to using the last hidden state of the Transformer as the representation and applying GAP for spatial squeeze—the model performs the worst across all cases. However, when $m$ and $n$ were set to values greater than 1, the model's performance significantly improved with minimal variation between them. This means that in practice, we only need to select blocks and frequency components greater than 1. This result is encouraging as it highlights the effectiveness of our approach while reducing the dependence on hyperparameter tuning, thereby demonstrating the ease of integration of S²Sent.

\begin{figure}[t]
\centering
\includegraphics[width=1\linewidth]{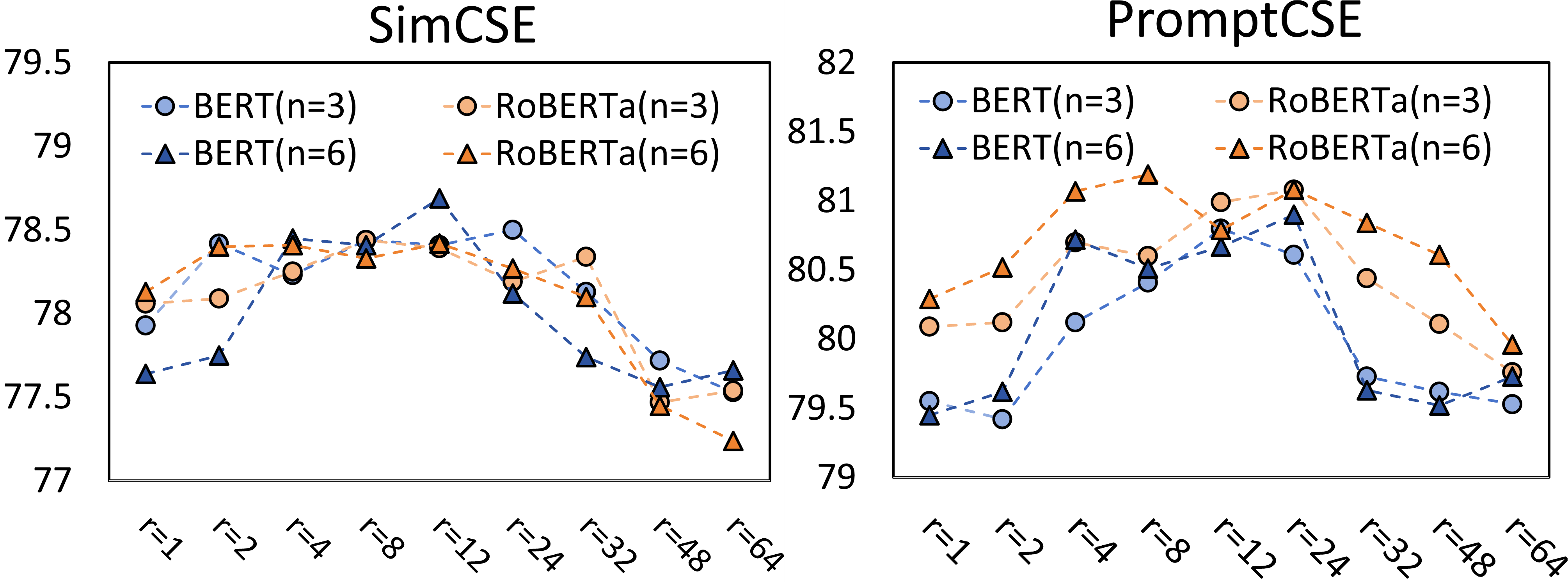}
\caption{The impact of $r$}
\label{fig.r}
\end{figure}

\subsubsection{Effect of Reduction ratio $r$}

%$r$作为Reduction ratio，它在S\textsuperscript{2}Sent的excitation stage构造了瓶颈层，大小与模型规模呈线性负相关，避免参数的冗余。我们对不同的r值进行了实验，Figure~\ref{fig.r}中的比较表明，baseline的性能随着$r$的增加先增后减。这可能是过小的$r$导致大参数量，从而产生过拟合。而过大的$r$导致的过低秩瓶颈会损害模块的表达能力。综合考虑，$r$=4--$r$=24均是一个可行的范畴。

%$r$ as the reduction ratio, constructs a bottleneck layer in the excitation stage. Its size is linearly negatively correlated with the model size to reduce parameter redundancy. We conducted experiments with different values of $r$, and the comparison in Figure~\ref{fig.r} indicates that the performance of the baseline initially increases and then decreases with increasing $r$. This could be because a too small $r$ leads to a large number of parameters, resulting in overfitting. On the other hand, an excessively large $r$ with a low-rank bottleneck can impair the expressive power of the module. Accordingly, $r$=4 to 24 are all feasible values.

The reduction ratio $r$ constructs a bottleneck layer in the excitation stage. Its size is linearly negatively correlated with the model size to reduce parameter redundancy. We conducted experiments with various values of $r$, and the comparison shown in Figure~\ref{fig.r} indicates that the performance of the baseline initially increases and then decreases as $r$ increases. This is likely because a very small $r$ results in a large number of parameters, leading to overfitting. Conversely, an excessively large $r$ can impair the module's expressive power due to an overly reduced bottleneck. Therefore, values of $r$ ranging from 4 to 24 are all considered feasible.

\subsection{Case Study}

\begin{figure}[t]
\centering
\includegraphics[width=1\linewidth]{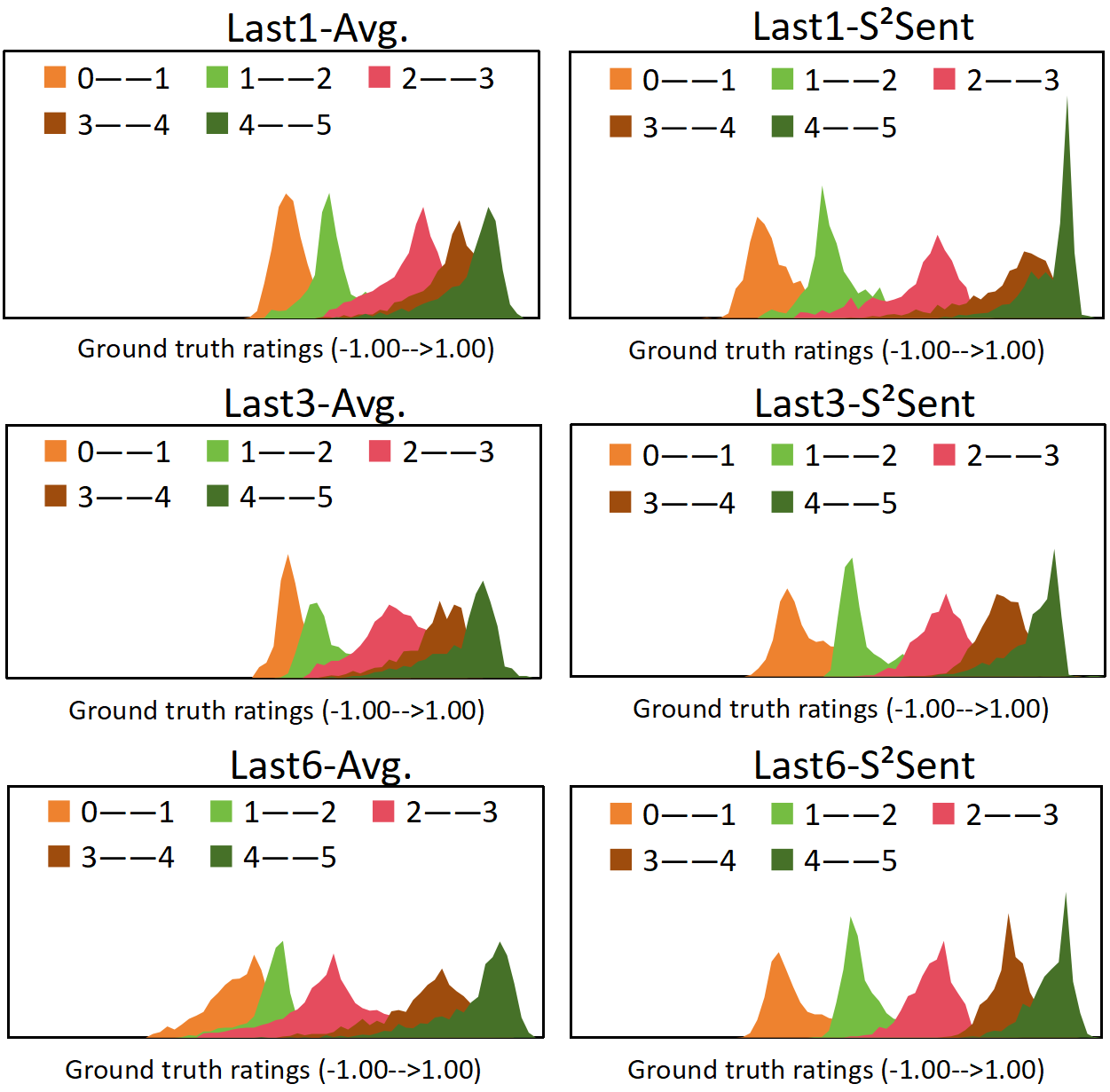}
\caption{Density plots of cosine similarities between sentence pairs in STS-B. Pairs are divided into 5 groups
based on ground truth ratings (higher means more similar) along the y-axis, and x-axis is the cosine similarity.}
\label{fig.des}
\end{figure}

%为了直接展示我们的方法在STS任务上的优势，我们在Figure~\ref{fig.des}展示了在BERT-SimCSE上cosine similarity distributions of STS-B pairs 和它们所对应的groundtruth human ratings group。我们主要比较了在基线上直接融合不同block（last1，3，6）的hidden states作为句子表示和采用S\textsuperscript{2}Sent优化句子表示下不同human rating组别的余弦分布。相较于直接融合不同block中hidden states作为sentence representation，引入S\textsuperscript{2}Sent的基线方法表现出更分散的分布，同时在语义相似的句子对上也保持了较低的方差，即更好地区分不同相似程度的句子对。

%To directly demonstrate the advantages of our method on the STS task, we present in Figure~\ref{fig.des} the cosine similarity distributions of STS-B sentence pairs using BERT-SimCSE and thier corresponding different groups of human ratings. Specifically, we compare the direct fusion of hidden states from different blocks (last 1, 3, 6) as sentence representations in the baseline approach with the utilization of S\textsuperscript{2}Sent to optimize sentence representations. In comparison to the direct fusion of hidden states from different blocks as sentence representations, the baselines incorporating S\textsuperscript{2}Sent exhibits a more scattered distribution while maintaining lower variance even for semantically similar sentence pairs. This indicates a better differentiation of sentence pairs with varying degrees of similarity.

To directly demonstrate the advantages of our method on the STS task, we present in Figure~\ref{fig.des} the cosine similarity distributions of STS-B sentence pairs using BERT-SimCSE and their corresponding human ratings. Specifically, we compare the direct fusion of hidden states from different blocks (last 1, 3, 6) as sentence representations in the baseline approach with the utilization of S\textsuperscript{2}Sent for optimizing sentence representations.
Compared to the direct fusion of hidden states from different blocks, the baselines incorporating S\textsuperscript{2}Sent exhibit a more scattered distribution of cosine similarities while maintaining lower variance, even for semantically similar sentence pairs. This indicates that S\textsuperscript{2}Sent provides better differentiation of sentence pairs with varying degrees of similarity.

\section{Conclusion}

%本文提出了一种创新的句子表示技术S\textsuperscript{2}Sent，它在Transformer-based encoder下游builds了一个参数化嵌套选择器，由空间选择（SS）和嵌套在SS中的频率选择（FS）组成。SS是一种选择性多尺度自适应融合，。FS是一种多光谱选择性空间squeeze。S\textsuperscript{2}Sent在疏导不同block的特征提取的同时，在下游尽可能避免语义信息的损失或冗余。不仅限于句子表示，S\textsuperscript{2}Sent的框架适用于所有基于stack或Inception的编码结构，这让我们看到了它在其他应用场景下的潜力。

We proposes an innovative sentence representation technique called S\textsuperscript{2}Sent, which builds a parameterized nested selector downstream of a Transformer-based encoder for cross block representation fusion. The nested selector consists of Spatial Selection (SS) and Frequency Selection (FS) nested within SS. SS is a selective multi-scale adaptive fusion technique, while FS is a multi-spectral selective spatial squeeze. S\textsuperscript{2}Sent enables effective sentence representation fusion across different blocks while minimizing the loss or redundancy of semantic information in downstream tasks. Beyond sentence representation, the framework of S\textsuperscript{2}Sent is applicable to stack-based or Inception-based encoding structures, revealing its potential in other application scenarios.

\section{Limitations}

%考虑到句子表征训练范式的差异，我们的S2Sent模块适用于基于transformer-based encoder的预训练模型，而不适用于基于生成式的LLM进行sentence embedding的训练。在基于生成式LLM sentence representation是否支持这种跨层表征动态融合是一个值得探索的可能性。

Considering the differences in sentence representation training paradigms, our S\textsuperscript{2}Sent is suitable for transformer-based encoder pre-trained models but not applicable to training sentence embeddings with generative LLMs. Whether generative LLM-based sentence representations can support such dynamic cross-layer representation fusion remains an open question worth exploring.

\bibliography{cv}
\appendix

\section{Contrastive Learning}

%我们在训练部分采用了~\cite{gao2021simcse}的方法，应用对比学习来微调句子表示。使用对比损失需要成对的例子，$\mathcal{B} = {(\bm{v}_d，\bm{v}_d^+)}$作为训练集中的一个batch，其中$\bm{v}_d$作为句子表示，$\bm{v}_d^+$是一个相关正例的表示（在语义上接近的）。在无监督训练时，$\bm{v}_d^+$作为$\bm{v}_d$的正例，batch中其他样本都视作负例。模型应该学习在让整理靠近样本且负例远离样本。我们使用批量采样的softmax来构造我们的对比损失，如公式~\ref{eq.unsup}所示。

We applied contrastive learning for training sentence embeddings. The contrastive loss requires pairs of examples, $\mathcal{B} = \{(\bm{v}_d, \bm{v}_d^+)\}$, as a batch in the training set, where $\bm{v}_d$ represents sentence representation \emph{after pooling} and $\bm{v}_d^+$ represents a representation of a relevant positive example (one that is semantically similar). During unsupervised training, $\bm{v}_d^+$ serves as the positive example for $\bm{v}_d$, while all other samples in the batch are treated as negatives. The model is trained to bring positives closer and push negatives away. We employed batch-sampled softmax to construct our contrastive loss, as formulated in Equation~\ref{eq.unsup}. Here, $\mathcal{B}$ represents a training batch, and $\tau$ is the softmax temperature. We use cosine similarity as the similarity function $\mathcal{F}_{\text{sim}}(\bm{v}_d,\bm{v}_d^+)=\frac{\bm{v}_d^\text{T} \cdot \bm{v}_d^+}{|\bm{v}_d||\bm{v}_d^+|}$.  

\vspace{-8pt}
\begin{equation}
    \mathcal{L}=-\log\frac{\exp(\frac{\mathcal{F}_{\text{sim}}(\bm{v}_d,\bm{v}_d^+)}{\tau})}{\sum_{k \in \mathcal{B}}\exp(\frac{\mathcal{F}_{\text{sim}}(\bm{v}_d,\bm{v}_k^+)}{\tau})}\label{eq.unsup}
\end{equation}

%其中我们用向量余弦作为相似度计算函数$\mathcal{F}_{\text{sim}}(\bm{v}_d,\bm{v}_d^+)=\frac{\bm{v}_d^T \dot \bm{v}_d^+}{|\bm{v}_d|| \bm{v}_d^+|}$。$\mathcal{B}$是一个训练batch，$\tau$是 softmax temperature。

% 值得注意的是，在将2D的句子表示池化成1D的表示时，一般有采用CLS的表示或者是所有token表示的平均值这两种选择，我们在这一部分讨论在引入S\textsuperscript{2}Sent时这两种池化对句向量的影响。Table~\ref{tab.pooling}讨论用SimCSE训练句向量情况下不同句embedding池化的影响，并报告STS的平均值。我们惊讶的发现不引入S\textsuperscript{2}Sent时，CLS和Avg.的结果无明显差异，当引入了S\textsuperscript{2}Sent时，采用CLS池化的句子embedding的表现有了令人惊讶的锐减。这是因为不同block的CLS表示代表了各层的语义信息，但不像Avg.池化和token的表示呈线性关系当通过S\textsuperscript{2}Sent加权融合后的CLS表示并失去了大部分的语义信息，失去了它本身的意义。所以我们强烈主张在对比学习中采用平均池化获得句子的1D embedding进行下游对比学习。

It is worth noting that when pooling 2D sentence representations into 1D representations, there are generally two options: using the [CLS] representation or the average of all token representations. In this section, we discuss the impact of these two pooling methods on sentence vectors when introducing S\textsuperscript{2}Sent. Table~\ref{tab.pooling} discusses the impact of different sentence embedding pooling on sentence vectors trained with SimCSE and reports the average of STS. We were surprised to find that without the introduction of S\textsuperscript{2}Sent, there was no significant difference between the results of [CLS] and Avg.. However, when S\textsuperscript{2}Sent was introduced, the performance of sentence embeddings using [CLS] pooling experienced a surprisingly sharp decline. This is because the [CLS] representation of different blocks captures the semantic information of each block, but unlike the Avg. pooling, the [CLS] representation does not have a linear relationship with the token representations when weighted and fused through S\textsuperscript{2}Sent, losing most of its semantic information and its original meaning. Therefore, we strongly advocate for the use of average pooling to obtain the 1D embedding of sentences for downstream contrastive learning.

% \usepackage[normalem]{ulem}
\begin{table}[h]
\centering
\begin{tabular}{ccc}
\hline
\textbf{Pooling} & \textbf{BERT}    & \textbf{+S\textsuperscript{2}Sent}   \\ \hline
CLS              & 76.21            & \textbf{66.88} \\ 
Avg.              & 76.22            & 78.50                 \\ \hline\hline
\textbf{Pooling} & \textbf{RoBERTa} & \textbf{+S\textsuperscript{2}Sent}   \\ \hline
CLS              & 76.65            & \textbf{61.17} \\ 
Avg.              & 76.46            & 78.48                \\ \hline
\end{tabular}
\caption{The impact of different sentence embedding pooling on the training of sentence embedding with SimCSE.}\label{tab.pooling}
\end{table}

\section{Transfer Tasks}

We also evaluated our model on the following transfer tasks: MR, CR, SUBJ, MPQA, SST-2, TREC, and MRPC. We followed the default configurations in SentEval. The results are shown in Table~\ref{tab.transfer}. It can be observed that after the introduction of S2Sent, the baseline showed significant improvements on most of the benchmarks.

\begin{table*}[t]
\centering
\begin{tabular}{lcccccccc}
\hline
Method                & MR             & CR             & SUBJ           & MPQA           & SST-2          & TREC           & MRPC           & Avg.           \\ \hline
SimCSE-BERT           & 81.01          & 86.19          & 94.45          & 89.28          & 85.10          & 89.40          & 74.51          & 85.71          \\ 
+S\textsuperscript{2}Sent    & 80.95          & \textbf{86.79} & \textbf{95.12} & \textbf{89.97} & \textbf{86.12} & \textbf{89.97} & \textbf{74.98} & \textbf{86.27} \\ \hline
PromptBERT            & 80.69          & 85.89          & 93.81          & 89.87          & 84.89          & 88.48          & 76.28          & 85.70          \\ 
+S\textsuperscript{2}Sent    & \textbf{81.45} & \textbf{86.64} & 93.91          & 89.91          & 84.77          & 89.52          & \textbf{76.99} & \textbf{86.17} \\ \hline
SimCSE-RoBERTa        & 81.15          & 87.14          & 93.29          & 86.14          & 86.50          & 84.43          & 73.28          & 84.56          \\ 
+S\textsuperscript{2}Sent & \textbf{82.01} & \textbf{87.89} & \textbf{93.96} & 85.91          & \textbf{86.94} & \textbf{85.23} & \textbf{74.02} & \textbf{85.14} \\ \hline
Prompt-RoBERTa        & 83.72          & 88.59          & 93.37          & 90.40          & 88.10          & 90.01          & 76.27          & 87.21          \\ 
+S\textsuperscript{2}Sent & 83.44          & \textbf{88.90} & 93.01          & \textbf{90.77} & \textbf{88.64} & 90.08          & 76.43          & 87.32          \\ \hline
\end{tabular}
\caption{Transfer task results of different sentence embedding models. The bolded values represent results that show significant improvement relative to the baseline (t-test).}\label{tab.transfer}
\end{table*}

\section{SS Motivation's Supplementary Explanation}

%我们在正文中解释了SS引入了多尺度自适应融合后，每个$\prod_{k=0}^n\frac{\partial \bm{u}^{(k)}}{\partial \bm{u}^{(k-1)}}$前的系数$\mathcal{K}=\bm{e}^{(n)}+\frac{\partial \bm{v}}{\partial \bm{U}}$与$n$有关。而传统的直接线性融合不同Transformer block的条件下，$\mathcal{K}$即为一个常数或者是与$n$无关的梯度值。我们将在这一部分分析线性融合下的梯度管理，即不为每个block分配自适应权重的情况。把每个Transformer的表示的聚合$bm{U}$与其特征嵌入进行元素级乘法，有$\bm{v}=\bm{e}*\bm{U}$。梯度流动链由$\bm{e}$和、$\bm{U}$的两条梯度传播共同决定，即$\frac{\partial \bm{v}}{\partial \bm{U}=\bm{e}$和$\frac{\partial\bm{v}}{\partial \bm{U}}=\bm{e}$。其中$\bm{e}$方向的梯度流动如公式~\ref{eq.e}所示。

We explained in the main text that after SS introduced multi-scale adaptive fusion, the coefficient in front of each \(\prod_{k=0}^n\frac{\partial \bm{u}^{(k)}}{\partial \bm{u}^{(k-1)}}\) is \(\mathcal{K}=\bm{e}^{(n)}+\frac{\partial \bm{v}}{\partial \bm{U}}\), which is related to \(n\). Under the condition of directly linearly fusing different Transformer blocks traditionally, \(\mathcal{K}\) is either a constant or a gradient value unrelated to \(n\). We will analyze gradient management under linear fusion in this part, that is, the case of not assigning adaptive weights to each block. The aggregation of each Transformer's representation \(\bm{U}\) is multiplied element-wise with its feature embedding, resulting in \(\bm{v}=\bm{e} * \bm{U}\). The gradient flow chain is jointly determined by the two gradient propagations of \(\bm{e}\) and \(\bm{U}\), that is, \(\frac{\partial \bm{v}}{\partial \bm{U}}=\bm{e}\) and \(\frac{\partial \bm{v}}{\partial \bm{e}}=\bm{U}\). The gradient flow in the direction of \(\bm{e}\) is shown in Equation~\ref{eq.e}.

\begin{equation}
\begin{aligned}
       &\frac{\partial \bm{v}}{\partial \bm{U}}|_{\rm w/o} =  \frac{\partial \bm{v}}{\partial \bm{e}} * \frac{\partial \bm{e}}{\partial \bm{s}} * \frac{\partial \bm{s}}{\partial \bm{v}}*\frac{\partial \bm{v}}{\partial \bm{U}} \label{eq.e} 
\end{aligned}
\end{equation}

Combining the gradient backpropagation of \(\bm{u}^{(n)}\rightarrow \bm{U}\), the calculation of the total gradient flow is shown in Equation~\ref{eq.grad1}. Where \(\bm{u}^{(n)} \rightarrow \bm{U}\) is a stacked linear connection, hence \(\frac{\partial \bm{x}}{\partial \bm{x}^{(n)}}\) is a constant. To simplify notation, we take it as 1 and omit it in the Equation~\ref{eq.grad1}.

\vspace{-8pt}
\begin{equation}
\begin{aligned}
        \frac{\partial \bm{v}}{\partial \bm{u}^{\rm init}}|_{\rm w/o}&=2*\frac{\partial \bm{v}}{\partial \bm{U}}*\sum_{n=0}^{N-1}\frac{\partial \bm{U}}{\partial \bm{u}^{(n)}}*\prod_{k=0}^n\frac{\partial \bm{u}^{(k)}}{\partial \bm{u}^{(k-1)}}\\&=2*\frac{\partial \bm{v}}{\partial \bm{U}}*\sum_{n=0}^{N-1}\prod_{k=0}^n\frac{\partial \bm{u}^{(k)}}{\partial \bm{u}^{(k-1)}}\label{eq.grad1}
\end{aligned}
\end{equation}

% %可以观察到，每个Transformer block进行特征提取的梯度流$\prod_{k=0}^n\frac{\partial \bm{u}^{(k)}}{\partial \bm{u}^{(k-1)}}$，以$2*\frac{\partial \bm{v}}{\partial \bm{U}}$的相同比例影响着总梯度流。这表明个Transformer block的参数以相同的程度进行更新。然而，这并不符合对不同尺度语义特征进行差异化梯度流管理的需求，导致Transformer中不同block的特征提取能力无法得到完全的释放。

It can be observed that the gradient flow for feature extraction in each Transformer block, \(\prod_{k=0}^n\frac{\partial \bm{u}^{(k)}}{\partial \bm{u}^{(k-1)}}\), affects the total gradient flow in the same proportion as \(2*\frac{\partial \bm{v}}{\partial \bm{U}}\). This indicates that the parameters of each Transformer block are updated to the same extent. However, this does not meet the requirement for differentiated gradient flow management of semantic features at different scales, resulting in the feature extraction capabilities of different blocks within the Transformer not being fully unleashed.

\end{document}